\documentclass[10pt, logo, twocolumn, copyright]{nvidiatechreport}

\usepackage{mdframed}
\usepackage[utf8]{inputenc} % allow utf-8 input
\usepackage[T1]{fontenc}    % use 8-bit T1 fonts

\usepackage{amsfonts}       % blackboard math symbols
\usepackage{nicefrac}       % compact symbols for 1/2, etc.
\usepackage{microtype}      % microtypography
\usepackage[dvipsnames]{xcolor}         % colors
\usepackage[dvipsnames]{xcolor}         % colors
\definecolor{mediumgray}{gray}{0.6}
\usepackage{multirow}
\usepackage{subcaption}
\usepackage{multirow}
\usepackage{multicol}
\usepackage{graphicx}
\usepackage[numbers]{natbib}
\usepackage{tabto}
\usepackage{xspace}
\usepackage{amsmath}
\usepackage{adjustbox}
\usepackage{enumitem}
\usepackage{wrapfig}
\usepackage{dblfloatfix}

\usepackage{footmisc}

\usepackage{float}

\usepackage{natbib}
\usepackage{hyperref}
\usepackage{url}
\usepackage{booktabs}
\usepackage{multirow}
\usepackage{enumitem}
\usepackage{adjustbox}
\usepackage{tabularx}
\usepackage{arydshln}
\usepackage{wrapfig}

\usepackage{multirow}
\usepackage{colortbl}
\usepackage{amsmath}
\usepackage{makecell}
\usepackage{hhline}
\usepackage{array}
\usepackage{diagbox}
\usepackage{graphicx}
\usepackage{adjustbox}

\usepackage{fp} % Required for floating-point calculations

\usepackage{xcolor}
\usepackage[subtle]{savetrees}

\begin{document}

\title{VILA-M3: Enhancing Vision-Language Models with Medical Expert Knowledge}

\author{Vishwesh Nath\textsuperscript{1} \qquad Wenqi Li\textsuperscript{1} \qquad Dong Yang\textsuperscript{1} \qquad Andriy Myronenko\textsuperscript{1} \qquad Mingxin Zheng\textsuperscript{1} \qquad Yao Lu\textsuperscript{1} \qquad Zhijian Liu\textsuperscript{1} \qquad Hongxu Yin\textsuperscript{1} \qquad Yucheng Tang\textsuperscript{1} \qquad Pengfei Guo\textsuperscript{1} \qquad Can Zhao\textsuperscript{1} \qquad Ziyue Xu\textsuperscript{1} \qquad Yufan He\textsuperscript{1} \qquad Yee Man Law\textsuperscript{2} \qquad Benjamin Simon\textsuperscript{3} \qquad Stephanie Harmon\textsuperscript{3} \qquad  Greg Heinrich\textsuperscript{1} \qquad Stephen Aylward\textsuperscript{1} \qquad 
Marc Edgar\textsuperscript{1} \qquad Michael Zephyr\textsuperscript{1} \qquad Song Han\textsuperscript{1} \qquad Pavlo Molchanov\textsuperscript{1} \qquad Baris Turkbey\textsuperscript{1} \qquad Holger Roth\textsuperscript{1,$\ddag$} \qquad 
Daguang Xu\textsuperscript{1,$\ddag$} \\~\\
\textsuperscript{1}NVIDIA \quad \textsuperscript{2}SingHealth \quad \textsuperscript{3}NIH  \\
%\textsuperscript{$\dag$}Equal contribution \quad
\textsuperscript{$\ddag$}Equal advisory}

\correspondingauthor{X}

\begin{abstract}
\textbf{Abstract:} \hspace{2pt}
%
% CVPR Submission Field TL;DR: Generic vision language models are not precise enough for medicine, expert models are necessary for improving the required precision
%
Generalist vision language models (VLMs) have made significant strides in computer vision, but they fall short in specialized fields like healthcare, where expert knowledge is essential. 
%In traditional computer vision tasks, creative or approximate answers may be acceptable, but in healthcare, precision is paramount.  
Current large multimodal models like Gemini and GPT-4o are insufficient for medical tasks due to their reliance on memorized internet knowledge rather than the nuanced expertise required in healthcare. Meanwhile, existing medical VLMs (e.g. Med-Gemini) often lack expert consultation as part of their design, and many rely on outdated, static datasets that were not created with modern, large deep learning models in mind. VLMs are usually trained in three stages: vision pre-training, vision-language pre-training, and instruction fine-tuning (IFT). IFT has been typically applied using a mixture of generic and healthcare data. In contrast, we propose that for medical VLMs, a fourth stage of specialized IFT is necessary, which focuses on medical data and includes information from domain expert models.
Domain expert models developed for medical use are crucial because they are specifically trained for certain clinical tasks, e.g. to detect tumors and classify abnormalities through segmentation and classification, which learn fine-grained features of medical data$-$features that are often too intricate for a VLM to capture effectively.
%especially in radiology
This paper introduces a new framework, VILA-M3, for medical VLMs that utilizes domain knowledge via expert models. 
%Domain expert models developed for medical use were trained to detect tumors and classify abnormalities (segmentation, classification), which learn fine-grained features of the data compared to coarse features that a VLM can typically process. 
We argue that generic VLM architectures alone are not viable for real-world clinical applications and on-demand usage of domain-specialized expert model knowledge is critical for advancing AI in healthcare. Through our experiments, we show an improved state-of-the-art (SOTA) performance with an average improvement of $\sim$9\% over the prior SOTA model Med-Gemini and $\sim$6\% over models trained on the specific tasks. Our approach emphasizes the importance of domain expertise in creating precise, reliable VLMs for medical applications.
%\yin{looks slightly long ;)}

\vspace{5pt}

\textbf{Links:} \hspace{1pt} \href{https://github.com/Project-MONAI/VLM}{Code} (GitHub) | \href{https://huggingface.co/MONAI}{Models} (Hugging Face) | \href{https://vila-m3-demo.monai.ngc.nvidia.com/}{Demo}

\vspace{10pt}

\end{abstract}

\maketitle

%This is the best paper. You can modify single/double column format with \textbf{twocolumn}. 

\section{Introduction}
\label{sec:intro}

\begin{figure*}[htb!]
    \centering
    \includegraphics[width=0.85\textwidth]{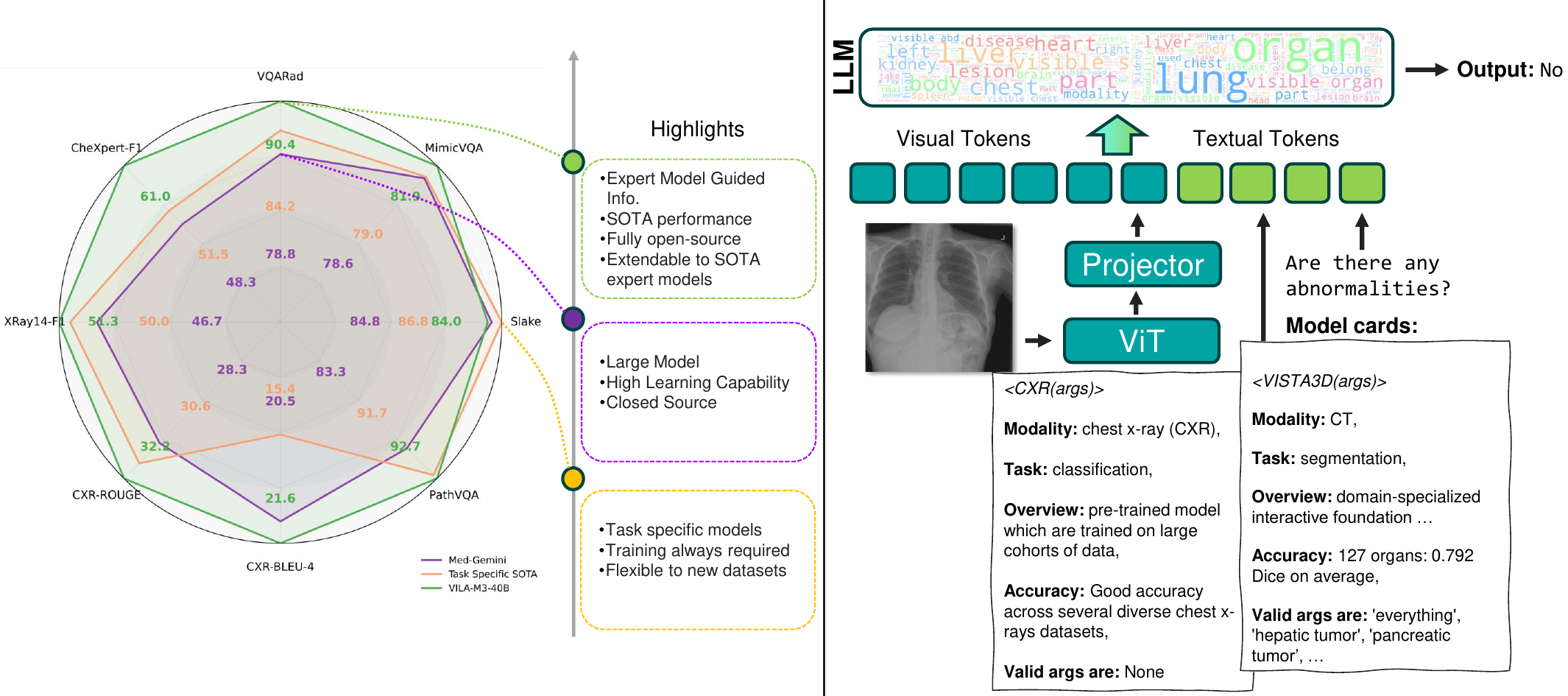}
    \caption{Left: Comparison of VILA-M3 with SOTA benchmarks such as Med-Gemini and task-specific SOTA models. VILA-M3-40B performance is shown in comparison. It can be observed that VILA-M3 provides a generalizable better performance for all datasets. Right: VILA-M3 architecture overview, the model aligns visual features using a projection layer with textual user prompts and model cards describing available ``expert models''.}%\yin{may merge with Fig. 2 and enhance style}
    \label{fig:radar_arch_plot}
\end{figure*}

%\yin{we can change hyperlink color to say nvgreen or calm colors}
Large language models (LLMs) are at the forefront of AI. Recent works~\cite{li2024llava,lin2024vila} have enabled LLMs to support visual inputs to expand on their use for many vision applications by transforming them into vision language models (VLM). 
While foundational VLMs developed for computer vision applications have achieved remarkable success in various general-purpose tasks, they struggle to meet the necessary precision required for medical tasks~\cite{moor2023med, tiu2022expert, bhayana2023performance}. VLMs trained on general tasks often lack specialized medical domain knowledge to interpret radiological images correctly \cite{hayden2024performance, bhayana2023performance, nisar2024d}. VLMs without specific medical training also often miss the subtle visual details that are critical for medical diagnosis. For example, a recent study on breast imaging \cite{cozzi2024bi} shows that generic models like GPT-4o (OpenAI) are not suitable for medical use cases. At the same time, there are healthcare-specific large VLMs such as Med-Gemini ($\sim$1.5 trillion parameters) \cite{yang2024advancing}, which are designed without consideration of expert model information and are likely to neglect finer visual features.  Current VLMs are well-adapted to connect coarse or dominating visual features with language. However, they often fail to recognize finer visual details (Fig. \ref{fig:qual_expert_seg}). There are multiple reasons for these deficits. A major reason is the limited availability of comprehensive and clinically relevant public datasets for medical vision language tasks. Most public datasets in medical imaging were released to address specific narrow AI tasks like classification, regression, and segmentation~\cite{li2023systematic}. At the same time, healthcare datasets incorporating natural language are limited to specific tasks such as visual question answering (VQA) and report generation, which do not cover the breadth of clinically relevant tasks. 

Over the years, medical imaging has shown successful applications of specialized narrow AIs that solve specific medical imaging tasks, such as semantic segmentation. These ``expert models'' have already been incorporated into clinical workflows, such as pre-operative planning, patient triage, and more. Many have received regulatory approval (e.g., FDA) and, therefore, incorporating the ``knowledge'' of existing expert models into the decision-making and generation process of VLMs, specifically in healthcare, should markedly improve the overall performance on clinically relevant tasks.

In this work, we propose a new VLM framework, \textbf{VILA-M3}, that addresses the unique challenges faced by general-purpose VLMs when applied to the medical domain by incorporating the domain-expert knowledge of existing segmentation and classification models. We evaluate our approach on several medical imaging benchmarks and show significant improvements over the state of the art (SOTA), see Fig.~\ref{fig:radar_arch_plot}.

\subsection{Key challenges}
Next, we summarize the key challenges that VILA-M3 aims to address.

\noindent\textbf{Limitations of Existing Medical Datasets:} The majority of publicly available medical datasets used for model training and evaluation are static and not tailored for training for large VLMs \cite{irvin2019chexpert, wang2017chestxray, he2020pathvqa, liu2021slake, lau2018dataset, ghaffari2019automated, baemimic}. The datasets were originally designed for earlier machine learning approaches, and they do not reflect the complexities required for modern deep learning models such as VLMs. This need has sparked the generation of newer datasets such as \cite{xie2024medtrinity, chen2024detecting, hu2024omnimedvqa}. In medicine, the constant need for expert validation and the evolving nature of healthcare data make static datasets insufficient for accurate model evaluation. In this work, we outline the need for more dynamic and adaptive approaches and show that on-demand usage of domain expert models can lead to improved model performance. The dataset limitation is one of the primary reasons we require and utilize expert model information in our framework. While large models like GPT-4o have been suggested as potential tools for evaluating model performance in medicine, their reliance on memorized information makes them unreliable for precision-critical tasks \cite{hayden2024performance, bhayana2023performance}.

%\textbf{The Need for Expert-Guided Evaluation Frameworks}:  There is a growing necessity for human-in-the-loop systems, where expert intervention plays a key role in assessing model outputs. Introducing an investigative questioning approach, validated by medical experts, can significantly improve the evaluation process for vision-language models in healthcare.

\noindent\textbf{Combining General Vision-Language Data with Healthcare Data:} Healthcare-specific datasets, while valuable, are not sufficient to fully train models that also require linguistic understanding. To overcome the aforementioned challenge, we utilize a four-stage training schema: pre-training vision encoder, pre-training VLM, instruction fine-tuning (IFT), and IFT with domain expert information. The two stages of IFT ensure that models can \textit{preserve} their language capabilities without sacrificing performance on medical benchmarks \cite{hartsock2024vision}. This approach not only enhances their overall performance but also addresses the unique challenges posed by medical language and its precision requirements.

\subsection{Contributions}
Motivated by the above challenges, we present the following key contributions of our work:
\begin{itemize}
    \item \textit{Expert Knowledge Integration}: We emphasize the need for integrating expert knowledge into medical VLMs to improve precision.

    \item \textit{Comprehensive capability}:
    %VILA as a VLM has been extended for the first time towards a medical application use-case to the best of our knowledge. 
    VILA-M3 is the first medical VLM that can tackle segmentation, classification, report generation, and VQA tasks in one framework. 
    
    %\item \textit{Dynamic Evaluation Framework}: We propose a dynamic evaluation method using expert-validated investigative questioning to replace outdated evaluations.
    
    \item \textit{Expert-guided instruction fine-tuning}: We preserve the VLMs language capabilities by introducing expert-guided IFT training on top of the built VLM training schemes.
    
    \item \textit{Hybrid 2D/3D Information Fusion}: We introduce effective integration with 2D and 3D expert models, enabling hybrid fusion of domain expert models that provide relevant 3D spatial information to enhance VLMs limited to 2D inputs.

    %\item VILA-M3 is a first step towards building chain of thought capabilities for VLMs by enabling the automated triggering of expert model inference to solve more complex tasks.

    \item \textit{Open-Source Module}: We provide an open-source module for data preparation, training, and model evaluation in medical VLM.

    %\item quantification of segmentation results with VLMs
\end{itemize}
\section{Related Work}
\label{sec:related_work}

%In this section we discuss the recent influential and works that inspired us to create VILA-M3.

%We examine the evolution of general vision-language models (VLMs) and highlight their limitations in medical applications that demand high precision and domain-specific expertise. We explore specialized medical vision models and expert systems that achieve state-of-the-art performance by leveraging expert knowledge from carefully curated datasets. Finally, we discuss the potential of agentic systems that utilize generative methods to develop powerful medical agents capable of integrating heterogeneous information and complex reasoning.

\begin{table*}[t!]
\centering
\caption{Performance of different models across various Visual Question Answering, Report Generation, and Classification benchmarks. Task-specific SOTA baselines are described in the experiments section \ref{sec:benchmarking}. Datasets where * is shown for Med-Gemini are either if a subset is used or pre-processing details have not been disclosed. (For VQARad, a particular subset was used and, pre-processing strategy for CXR dataset is undisclosed.)
The highest score for each task is highlighted in bold, while the second-best is indicated with an underscore.}
% \yin{we can merge params into names, and adjust width of column to make lines single lines. bold ranking first, and underline second best. may need to explain forgetting amid why are we so much better, not due to overfitting etc. This as main table, needs more work.}
\resizebox{\textwidth}{!}{%
\scriptsize % Set font size to scriptsize
\setlength{\tabcolsep}{0.5pt} % Adjust column separation to save space
\begin{tabular}{c | cccc ccc cc c}
\toprule
\rowcolor{lightgray}
\multicolumn{1}{c|}{\textbf{Model Type}} & \multicolumn{4}{c}{\textbf{Visual Question Answering}} & \multicolumn{2}{c}{\textbf{Report Generation}} & \multicolumn{2}{c}{\textbf{Classification}} & \multicolumn{1}{c}{\textbf{}} \\ \midrule
%\rowcolor{lightgray}
\makecell{\textbf{Datasets}} & \textbf{Rad} & \textbf{MIMIC} & \textbf{SLAKE} & \textbf{Path} & 
\makecell{\textbf{CXR (Exp.)}} & 
\makecell{\textbf{CXR (Exp.)}} & 
\makecell{\textbf{X-ray14 (Exp.)}} & 
\makecell{\textbf{CheXpert (Exp.)}} & 
\makecell{\textbf{  }\\\textbf{  }} \\ \midrule
%\rowcolor{lightgray}
\multicolumn{1}{c|}{\textbf{Metric}} & 
\textbf{Acc.} & 
\textbf{Acc.} & 
\textbf{Acc.} & 
\textbf{Acc.} & 
\textbf{BLEU-4} & 
\textbf{ROUGE} & 
\textbf{F1} & 
\textbf{F1} &
\textbf{Tot. Avg.}\\ \midrule
\makecell{\textbf{Med-Gemini (1.5T)}} & 78.8* & 78.6 & \underline{84.8} & 83.3 & 20.5* & 28.3* & 46.7 & 48.3 & 55.7 \\ 
\makecell{\textbf{Task Spfc. SOTA}} & 84.2 & - & \textbf{86.8} & \underline{91.7} & 15.4 & 30.6 & 50.0 & 51.5 & 58.6 \\ 
\makecell{\textbf{VILA-M3 (3B)}} & 78.2 & \underline{82.4} & 79.8 & 87.9 & 20.2 & 31.7 & \underline{51.3} & 60.8 & 61.5  \\ 
\makecell{\textbf{VILA-M3 (8B)}} & \underline{84.7} & 82.1 & 82.7 & 91.0 & 21.1 & 32.0 & 48.9 & \textbf{61.6} & 63.0 \\ 
\makecell{\textbf{VILA-M3 (13B)}} & 80.5 & \textbf{86.4} & 83.2 & 91.0 & \underline{21.6} & \underline{32.1} & 51.2 & \underline{61.5} & \underline{63.4} \\
\makecell{\textbf{VILA-M3 (40B)}} & \textbf{90.4} & 81.9 & 84.0 & \textbf{92.7} & \textbf{21.6} & \textbf{32.2} & \textbf{51.3} & 61.0 & \textbf{64.3} \\
\bottomrule
\end{tabular}}
\label{tab:mainbenchmark_table}
\end{table*}

\noindent\textbf{Vision-language models for medicine:} VQA tasks were one of the first that combined text and images directly in a single task~\cite{antol2015vqa}. These were later extended to FLAN-style visual instruction tuning tasks~\cite{wei2021finetuned,xu2024vision}. 
General-purpose large multi-modal models such as GPT-4~\cite{achiam2023gpt} and Gemini~\cite{team2023gemini} based on transformers~\cite{vaswani2017attention} have demonstrated great potential towards building intelligent conversational assistants built on combined textual and vision datasets. When applied in the biomedical domain, the models present promising capabilities of understanding the basic concepts. However, handling complex domain-specific tasks with high accuracy and fine granularity remains both challenging and demanding~\cite{yang2024advancing, li2024llava, saab2024capabilities, moor2023med}. Pre-training and instruction tuning is important for aligning the representations across modalities and enhancing the reasoning capabilities~\cite{lin2024vila, yang2024advancing}.

%They represented the early steps for building VLMs with conversational capabilities. 
Concurrently, there have been multiple methods targeting large VLMs for medicine in different aspects. Med-Gemini~\cite{yang2024advancing} boasts a 1.5 Trillion parameter model and includes non-vison tasks like genomics and medical exams. LLaVa-Med~\cite{li2024llava}, based on the popular LLaVa architecture~\cite{liu2024visual}, focused on using academic medical datasets, while Med-Flamingo \cite{moor2023med} is an extension of Flamingo \cite{alayrac2022flamingo} that introduced a novel mechanism for combining vision and text. BiomedParse allows promptable segmentations to be performed by the VLM itself~\cite{zhao2024biomedparse} but does not generalize to more complex tasks like report generation. While promising first steps towards VLMs in medicine, these works highlight the limitations in precision and lack of domain expertise when applied to medical tasks, compared to traditional computer vision applications on narrow tasks such as classification and segmentation, where traditional CNNs often outperform their VLM counterparts.
    
\noindent\textbf{Medical vision models and expert systems:} 
Many models have been developed to address domain- and task-specific challenges. They achieve state-of-the-art (SOTA) performances in certain areas and often focus on expert knowledge learned from carefully curated datasets. For example, Myronenko et al.~\cite{myronenko20193d} created accurate lesion segmentation models for multi-modal brain MRI, He et al.~\cite{he2024vista3d} studied 127 common types of human anatomical structures segmentation from tens of thousands of expert annotated CT examples. Several chest X-ray classification models have been developed by Cohen et al.~\cite{Cohen2022xrv} on a wide range of publicly available datasets. Integrating domain knowledge and guided reasoning while maintaining modular flexible system design can be a plausible approach to leverage the previous successes~\cite{wei2024mc,nisar2024d}.

\begin{figure*}[h]
    \centering
    \includegraphics[width=0.85\textwidth]{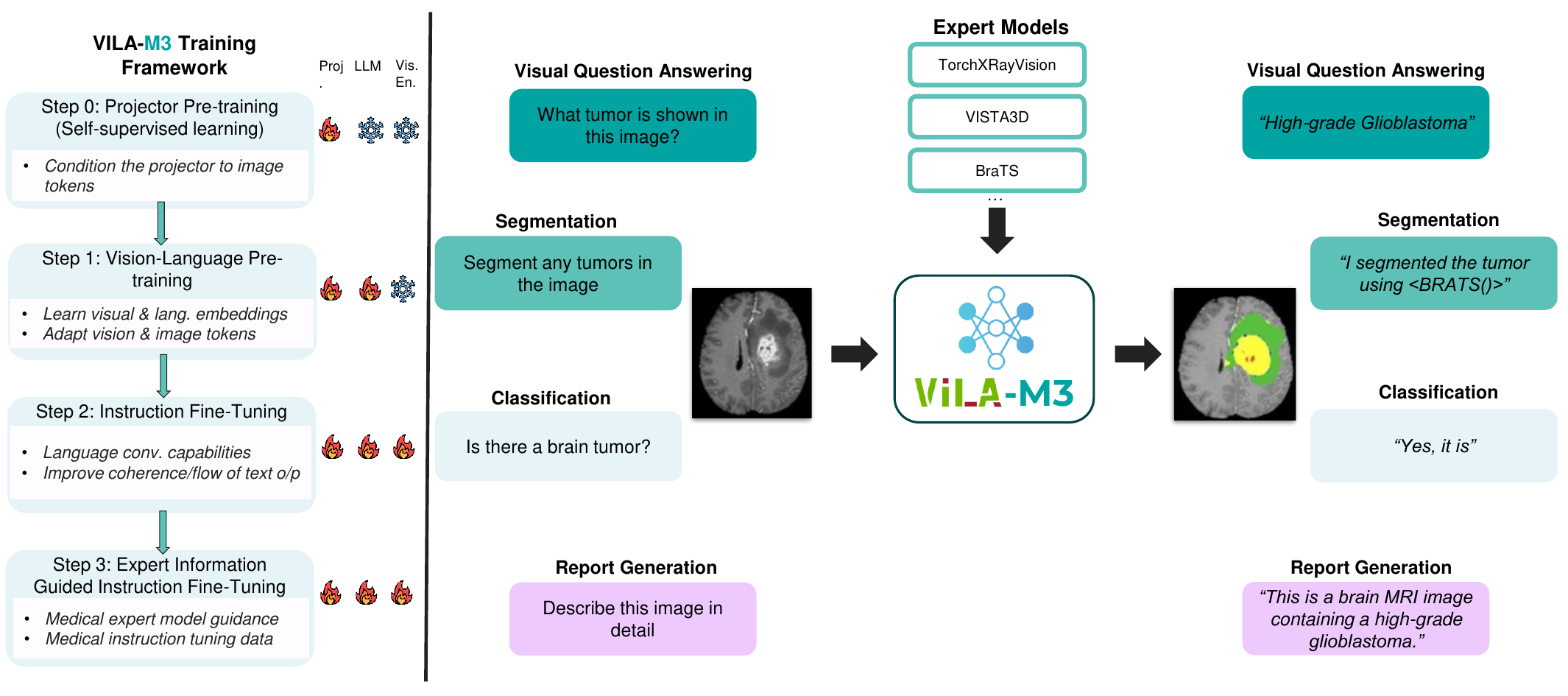}
    \caption{VILA-M3 possesses the capability to support a diverse range of tasks, including visual question answering, classification, and report generation.  Segmentation tasks are performed by suitable ``expert models'', such as the BraTS brain tumor segmentation model for multimodal MRI.
    }
    %\yin{given lots of space left, we may somehow enlarge fonts}}
    \label{fig:overview}
\end{figure*}

\noindent\textbf{Evaluation and benchmarking in healthcare AI:} Evaluation for vision-language models is a challenging task and is heavily dependent upon the type of task that is being evaluated. While it is straightforward to evaluate close-ended tasks such as classification, segmentation, or object detection as the well-established metrics can be directly used (e.g., F1, Dice, accuracy, etc.), it is not well-established how to evaluate more open-ended tasks where the answers have a wide range for variations. For example, open-ended questions in VQA and report generation require the evaluation of the correctness of entire sentences or paragraphs, which is challenging~\cite{ging2024open}. Many prior works attempted to evaluate the open-ended weaker VLM output by a stronger VLM. For instance, ChatGPT-4o is being used as a judge to evaluate the answer \cite{liu2024visual, li2024llava, wu2023towards}. %In contrast, we address the question \textit{``is the larger VLM a true oracle that can detect the accurate answer of a weaker VLM?"}
%There are numerous works that also compared the output of two or more VLMs via a stronger VLM to assess their performance \cite{liu2024visual, li2024llava, wu2023towards}. 
While using stronger VLMs as judges might be acceptable for computer vision tasks, this method of evaluation is not satisfactory for healthcare, where a factual level of precision is required, and patient outcomes depend on the results.
Our evaluation shows that the capabilities of off-the-shelf VLMs like ChatGPT-4o are too limited to judge the outputs of medical VLMs directly.
%
%Motivated by the above, we have included an expert radiologist opinion that covers the aspect of `what is useful' from the medical VLM's (Fig. \ref{fig:qual_report}).

\noindent\textbf{Agentic systems:} The flexibility of generative AI provides new possibilities for developing powerful medical agents that can integrate heterogeneous information and generic reasoning. Most previous works focus on extending large language models in a medical context~\cite{tang2023medagents,yang2024drhouse,schmidgall2024agentclinic,li2024agent,wei2024medco}. Very recently, Hoopes et al.~\cite{hoopes2024voxelprompt} developed a code-predicting agent framework that can integrate an array of external systems and APIs for a set of neuroimaging tasks including brain region analysis. Li et al.~\cite{li2024mmedagent} proposed a multi-modal large language model invoking task-specific tools and aggregating results from them.
\section{Method}

The VILA framework~\cite{lin2024vila} is suitable for pre-training vision and language models as well as instruction fine-tuning (IFT) them with domain-specific datasets. VILA-M3, in particular, leverages domain-expert model insights alongside the versatile capabilities of VILA to capture the nuances of radiological language and the intricate details of medical images, enabling more precise image-text fusion. 
%The VILA (Vision-Language Alignment) framework was designed to enhance the understanding of visual and text data by learning a joint representation space \cite{li2024llava, lin2024vila}. VILA was built upon the Llava framework \cite{li2024llava} but scales up the VLM pre-training by supporting higher-resolution input images and larger LLMs~\cite{lin2024vila}. 
%We employ data augmentations to diversify training inputs and include expert data to further enhance the overall performance of the model.
%VILA is a well-rounded technique that is broadly applicable to multiple data domains. 
%Our primary goal is to enable the model to correlate complex image patterns with associated text descriptions specifically for medicine, facilitating tasks such as visual question answering (VQA), zero-shot image classification, retrieval, report generation, and even expert model feedback for a more well-reasoned output. 
It enhances tasks such as visual question answering (VQA), zero-shot image classification, image retrieval, and report generation, and it even incorporates expert model feedback to produce more accurate and well-reasoned outputs.
VILA is particularly applicable for medical AI, where accurate association between visual data and textual information is critical for clinical decision-making. We further enhance it by including expert model feedback that can be triggered on demand. %and the information generated by them.

\subsection{Training Framework}

The standard approach to train VLMs like VILA is to use a corpus of visual-text data where text based conversations are associated with an image. The image token is randomly inserted in between the text or before/after the text (Fig.\ref{fig:radar_arch_plot}). As a common practice that the inclusion of text-only data is helpful towards preserving the language capabilities of VLM, we included a small corpus of medical text data \cite{jin2021disease} to specifically prevent degradation of the LLM capabilities as the model gets conditioned towards medical images.

%\subsection{Model Architecture}

VILA is based on auto-regressive multi-modal LLMs,  where images are tokenized into visual tokens, concatenated with text tokens, and fed into language models, effectively treating visual input as a foreign language (Fig.\ref{fig:radar_arch_plot}). Therefore, a projector layer is needed to bridge the token embeddings from the image and text modalities. While different projector architectures are possible, a simple linear layer encourages the LLM to generalize better~\cite{lin2024vila}. This approach naturally extends text-only LLMs by incorporating visual embeddings, allowing them to handle interleaved image-text inputs and generate text outputs. There are other frameworks available that combine vision and language, such as cross-attention based \cite{moor2023med, alayrac2022flamingo}, but auto-regressive approaches are gaining in popularity due to their flexibility in supporting different vision encoder and LLM backbones. Therefore, we chose VILA as it is the current SOTA for traditional computer vision and is an open-source framework. VILA models are typically trained in three training stages: projector pre-training, visual language pre-training, and IFT. VILA-M3 builds on the last stage of visual instruction tuning and expands it with the inclusion of information from domain expert models. While VILA has enough flexibility to allow for separate fine-tuning of the vision encoder, the projector, and the LLM, we fine-tune all as it is necessary to customize both vision, projector alignment, and LLM for the medical domain as shown in Fig.~\ref{fig:overview}.

\begin{figure*}[h!]
    \centering
    \includegraphics[trim={0 2em 0 0},clip,width=0.8\textwidth]{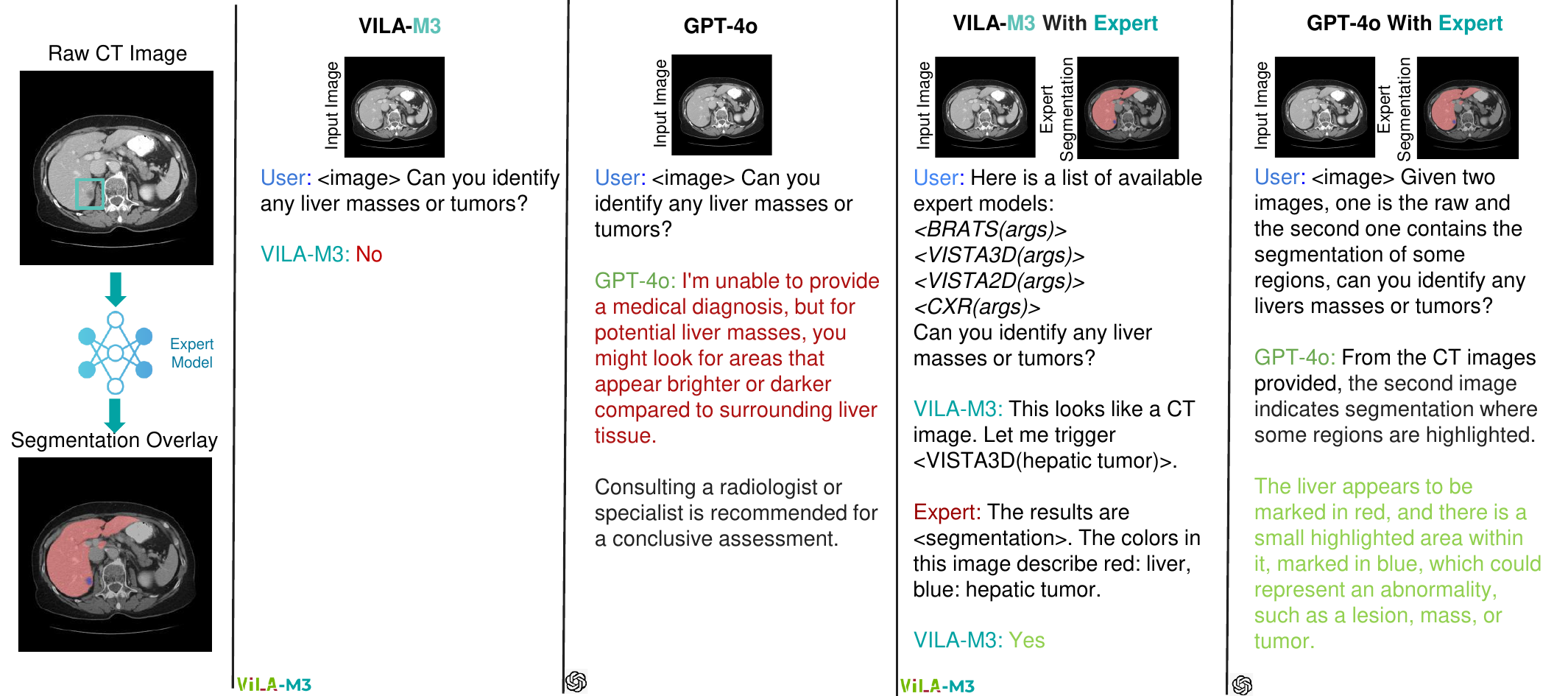}
    \caption{Feedback of segmentation results can improve the quality of responses received from VLMs. This observation holds true for both VILA-M3 and GPT-4o. The models without expert segmentation fail to detect the tumor, unlike the models with access to expert model segmentation. The blue annotation box shows the marked tumor location, traditional VLM’s cannot capture such fine features unless guided by expert outcomes.}
    \label{fig:qual_expert_seg}
\end{figure*}

%\textit{LLM:}
%\textit{Vision Encoder:}

\subsection{Expert Models Triggering \& Feedback}
In this work, we use open-source expert models for common medical imaging tasks. VILA-M3 learns to predict a keyword and argument, e.g., \texttt{<VISTA3D(hepatic tumor)>} when it wants to trigger a suitable expert model. The list of available expert models is fed as a system prompt to the model (see Fig.~\ref{fig:overview}).

\noindent\textbf{Used Expert Models:}
For volumetric segmentation in CT images, we utilize \textit{VISTA3D}%\footnote{\url{https://github.com/Project-MONAI/model-zoo/tree/dev/models/vista3d}}
~\cite{he2024vista3d}. VISTA3D is a domain-specialized interactive foundation model developed for segmenting and annotating human anatomies with precision. It achieves a highly accurate segmentation of 127 classes, including challenging tumors. We include several organ or anatomy subgroups that can be chosen by VILA-M3 using different arguments when predicting the model keyword. Valid arguments are `everything' (i.e., all 127 available organ classes), `hepatic tumor', `pancreatic tumor', `lung tumor', etc. These arguments are described in model cards available to VILA-M3 as context at training and inference time (see Fig.~\ref{fig:radar_arch_plot}). Note that when triggered, VISTA3D can segment the full volumetric scan even though VILA-M3 only processes 2D inputs.

For MRI imaging, we focus on tumor segmentation in multimodal brain MRI. The \textit{MONAI BRATS model}%\footnote{\url{https://github.com/Project-MONAI/model-zoo/tree/dev/models/brats_mri_segmentation}}
~\cite{myronenko20193d} consists of a pre-trained model for volumetric (3D) segmentation of brain tumor sub-regions from four MRI modalities (T1-weighted, T1-weighted with gadolinium contrast-enhancement, T2-weighted, and FLAIR) based on BraTS 2018 data and achieves good performance~\cite{menze2014multimodal}. %The model's average accuracy on tumor core, whole tumor, and enhancing tumor is 0.856, 0.903, and 0.791, respectively.
   
The majority of the data sets used for training contain chest X-rays (CXRs), which are used for many routine diagnostic clinical tasks. In this work, we utilize a model ensemble of CNN-based classification models from \textit{TorchXRayVision}%\footnote{\url{https://github.com/mlmed/torchxrayvision}}
~\cite{Cohen2022xrv}. These models are trained on large cohorts of data and provide good accuracy across several diverse CXR datasets.

\noindent\textit{Expert Model Feedback:}
When VILA-M3 selects and activates an appropriate expert model from the available model cards, the result is returned to VILA-M3 as another user prompt, formatted in a conversational style. For a segmentation result, we re-tokenize the generated mask and provide a textual description that explains the significance of the segmentation results, overlaid on the original image(see Fig.~\ref{fig:qual_expert_seg}). In the case of classification, we fed back the result as a list of yes/no statements for the likelihood that the image contains a certain disease (i.e., a list of 18 diseases using the TorchXRayVision ensemble).

\subsection{Datasets}

Multiple publicly available datasets were utilized for this study. We present a brief overview of them and the processing that we applied to effectively utilize them for training with the VILA framework. All training data is detailed in Supp. material. %can be observed in Table. \ref{tab:balanced_training_dataset_stats}

\noindent\textbf{VQA:}
In this study, we utilize the RadVQA, SLAKE, and PathVQA datasets to evaluate visual question answering (VQA) performance in medical imaging \cite{he2020pathvqa, liu2021slake, lau2018dataset}. The SLAKE dataset comprises 642 medical images across modalities such as CXR, CT, and MRI, accompanied by 14,028 question-answer pairs covering radiology across various anatomical regions. From the official dataset splits, we used 4,919 training, 1,053 validation, and 1,061 test examples, focusing on diverse question formats like open-ended and yes/no. The PathVQA dataset includes 4,289 pathology images paired with 32,632 question-answer pairs, divided into 19,654 for training, 6,259 for validation, and 6,719 for testing. Both datasets probe a wide range of visual and clinical attributes, including abnormalities, organ identification, and diagnostic information. Our analysis focuses on using these datasets to enhance the performance of medical vision language models. We utilized both the training and validation sets from all the VQA datasets as part of the model training, due to the huge compute cost of training VLM's standard validation techniques are quite compute heavy and also redundant.

\noindent\textbf{Classification:}
We adopted two publicly available high-quality datasets of abnormalities in CXR images as a benchmark for the model's visual classification capability. One is the subset of {ChestX-ray14}~\cite{wang2017chestxray} curated by Majkowska and Mittal et al.~\cite{doi:10.1148/radiol.2019191293}, consisted of 1,962 images after an adjudication process of four radiologists. For this dataset, we follow~\cite{yang2024advancing} and focus on the classification of three conditions -- lung opacity, pneumothorax, and fracture.
The other dataset is a subset of {CheXpert}~\cite{irvin2019chexpert} created by Saporta et al.~\cite{saporta2022benchmarking} on the labeling of five pathological conditions, namely atelectasis, cardiomegaly, consolidation, edema, and pleural effusion.

\noindent\textbf{Report Generation:}
For report generation, we adopt the MIMIC-CXR\footnote{v2.1.0} dataset~\cite{johnson2019mimiccxr,johnson2019mimic1cxrjpg}, which provides over 377,110 CXR images in JPG format, along with structured labels extracted from 227,827 associated free-text radiology reports.
%The images were originally in DICOM format and later converted to JPG for improved accessibility, using standardized image processing methods to adjust pixel values and contrast for optimal visual consistency.
Each image is linked to the corresponding radiology report, which contains the radiologists' findings and impressions.
Labels for the dataset were derived using automated tools CheXpert~\cite{irvin2019chexpert} and NegBio~\cite{peng2018negbio}, which extract and classify medical conditions (e.g., ``pneumonia,'' ``edema,'' ``pleural effusion'') as positive, negative, uncertain, or not mentioned based on the language of the reports.
Additionally, radiologists manually annotate a portion of the report, providing 14 categories of findings for model evaluation and validation.
The dataset is divided into training, validation, and testing parts, each with separate CSV files to support reproducible research.
Due to suboptimal report quality in the initial datasets, we applied advanced Large Language Models (LLMs) to refine text reports through targeted noise reduction and finding extraction.
This process significantly improved model accuracy and reliability.
Additional methodological details are available in the supplementary materials.

\noindent\textbf{Expert Datasets:}
To create training data that triggers expert models for tasks like segmentation and classification, we leverage existing datasets by running expert model inference. For instance, we can segment existing CT datasets with the VISTA3D model, using the inference results to build training conversations tailored for VILA-M3. Similarly, with a brain MRI dataset, we can run the BRATS segmentation model and then utilize the output to generate structured training conversations for training VILA-M3. This approach facilitates the generation of context-specific dialogues that enhance model alignment with expert-level tasks for medical imaging. Further, we extend a subset of the existing VQA and report generation tasks by adding the model cards to the message context and adding trigger statements for the corresponding CXR expert model as illustrated in Fig.~\ref{fig:radar_arch_plot}.
\section{Experiments \& Results}
\label{section:exp}
%\begin{table}[htbp]
%    \scriptsize
%    \centering
%    \caption{Fits to $L(N, S)$ of the loss scaling law defined in~\cite{kaplan2020scaling}. These parameters are visualized in Fig.~\ref{fig:loss_scaling_law}.}
%    \begin{tabular}{c|cccc}
%        \hline
%        \rowcolor{lightgray}
%        \textbf{Parameter} & $\alpha_N$ & $\alpha_S$ & $N_c$ & $S_c$ \\
%        \hline
%        \textbf{Value} & 0.78 & 1.09 & $1.50 \times 10^{8}$ & $3.92 \times 10^{2}$ \\
%    \end{tabular}
%    \label{scaling_law_paramters}
%\end{table}

\subsection{Model Architecture Hyper-parameters}
We use the image resolution in the visual encoder at 384$\times$384 using OpenAI's CLIP-L, enabling the model to capture more fine-grained visual details. For the LLM backbones, we use Vicuna (based on Llama-2 \cite{touvron2023llama}) for 3 Billion and 13 Billion parameter checkpoints, for 8 Billion, we utilize Llama-3 \cite{dubey2024llama} as the backbone and the 40 Billion uses Yi-34B as the LLM backbone based on Yi \cite{young2024yi} and uses a 448$\times$448 vision encoder. The LLM choices depend on the configurations used for the VILA checkpoints pre-trained on natural images.

\subsection{Implementation Details} \label{sec:training_hyperparameters}
\noindent\textbf{Training Hyper-parameters:} To ensure a fair comparison for all VILA model size variants, we kept the same training hyper-parameters for each size. A consistent batch size of 16 was maintained at the device (per GPU) level for training and 4 for evaluation. Gradient accumulation was adjusted accordingly to ensure the same overall effective batch size in multi-node training. The learning rate was initialized at 2e-5, using a cosine learning rate schedule with a warm-up ratio of 0.03. Weight decay was set to zero to avoid over-regularization, and all the models utilized bf16 mixed precision, allowing efficient training with a reduced memory footprint. 
Gradient check-pointing was enabled to further optimize memory usage. The LLM backbone, projector, and vision encoder were kept unfrozen as that is the best setting to adapt for new data as per \cite{lin2024vila} to fine-tune the entire model. 
%Models were saved or checkpointed at every 100 steps.

\noindent\textbf{Training Cost:} To train these models we used A100's with 80GB of memory. For the 3, 8, and 13 billion parameter models, we utilized four nodes (x8 A100) in parallel using 32 GPUs in total. For the 40B model, we utilized sixteen nodes, utilizing a total of 128 GPUs in parallel (Table~\ref{tab:training_compute_cost}). More information regarding inference computational cost can be found in the supplementary material.

\noindent\textbf{Inference Cost:}
With the open source release\footnote{Github ~\url{https://github.com/Project-MONAI/VLM}} the computational footprints are as follows: The CXR expert dynamically loads various TorchXrayVision models and performs ensemble predictions. The memory requirement is circa 1.5GB in total.
The VISTA3D expert model dynamically loads the VISTA3D model to segment a 3D-CT volume. The memory requirement is roughly 12GB, and peak memory usage can be higher, depending on the input size of the 3D volume.

\begin{table}[h!]
    \scriptsize
    \centering
    \caption{Training computational costs for different model sizes.}
    \resizebox{\columnwidth}{!}{%
    \begin{tabular}{c|ccc|c}
        \toprule
        \rowcolor{lightgray}
        \#Params & Nodes & Tot. GPUs & Epochs & \textit{Train Time} \\
        \midrule
        VILA-M3-3B & 4 & 32 & 2 & $\sim$ 5.5 hrs \\
        VILA-M3-8B & 4 & 32 & 2 & $\sim$ 11.0 hrs \\
        VILA-M3-13B & 4 & 32 & 2 & $\sim$ 19.5 hrs \\
        VILA-M3-40B & 16 & 128 & 2 & $\sim$21 hrs \\
        \bottomrule
    \end{tabular}}
    \label{tab:training_compute_cost}
\end{table}

\begin{table*}[t]
\centering
\caption{Performance comparison of VILA-M3 and VILA-1.5 models across traditional vision language model benchmarks. VQAv2 is testdev split and VizWiz is test split. Avg is calculated by averaging all the benchmark scores (note that MME is divided by 20).}
\resizebox{\textwidth}{!}{%
\scriptsize
\begin{tabular}{llr| rrrrrrrrrrrr}
\hline
\rowcolor{lightgray}
Model & LLM & Params & VQAv2 & GQA & VizWiz & SQA-I & VQA-T & POPE & MME & MMB & MMB-CN & SEED &  MM-Vet & Avg \\
%& & & (test) & & (test) & & & & & & & & bench & \\
\hline
VILA1.5 & Sheared LLaMa & 3B & 80.4 & 61.5 & 53.5 & 69.0 & 60.4 & 85.9 & 1442.4 & 63.4 & 52.7 & 60.9  & 35.4  & 63.2 \\
VILA-M3 & Sheared LLaMa & 3B & 73.9 & 56.0 & 38.4 & 49.5 & 53.3 & 84.0 & 1381.0 & 60.6 & 51.5 & 59.2  & 23.5 & 56.2\\
\hline
VILA1.5 & Llama-3 & 8B & 83.0 & 63.5 & 63.2 & 82.0 & 68.5 & 85.6 & 1634.9 & 75.3 & 69.9 & 66.4  & 43.2 & 71.1\\
VILA-M3 & Llama-3 & 8B & 70.4 & 53.2 & 38.4 & 72.3 & 51.2 & 80.8 & 1530.0 & 68.8 & 63.3 & 63.1  & 28.2 & 60.5\\
\hline
VILA1.5 & Vicuna-V1 & 13B & 82.8 & 64.3 & 62.6 & 80.1 & 65.0 & 86.3 & 1569.5 & 74.9 & 66.3 & 65.1  & 44.3 & 70.0\\
VILA-M3 & Vicuna-V1 & 13B & 78.4 & 61.2 & 55.9 & 74.1 & 60.2 & 83.7 & 1531.0 & 71.3 & 65.5 & 64.8  & 36.6 & 66.2\\
\hline
VILA1.5 & Yi-34B & 40B & 84.3 & 64.6 & 62.2 & 87.2 & 73.6 & 87.3 & 1726.8 & 82.4 & 80.2 & 69.1  & 53.0 & 75.5\\
VILA-M3 & Yi-34B & 40B & 70.2 & 50.7 & 41.3 & 0.0 & 53.0 & 82.9 & 1548.5 & 58.9 & 55.6 & 56.8  & 27.4 & 52.2\\
\hline
\end{tabular}}
\label{tab:vila_benchmark}
\end{table*}

\begin{table*}[htbp]
\centering
\caption{Performance comparison of various models on report generation utilizing the test set from the MIMIC-CXR-JPG Database~\cite{johnson2019mimiccxr,johnson2019mimic1cxrjpg}.}
% \scriptsize
\footnotesize
% \small
\begin{tabular}{lr|ccc|ccc}
\hline
\rowcolor{lightgray}
\multicolumn{2}{c|}{Model} & \multicolumn{3}{c|}{MIMIC-CXR w/o Expert} & \multicolumn{3}{c}{MIMIC-CXR with Expert} \\ \hline
\rowcolor{lightgray}
Type          & Params  & BLEU-4 (↑)    & ROUGE (↑)   & GREEN (↑)   & BLEU-4 (↑)    & ROUGE (↑)   & GREEN (↑)   \\ \hline
Med-Gemini~\cite{yang2024advancing}   & 1.5T       & 20.5          & 28.3        & -           & 20.5          & 28.3        & -           \\
Llava-Med~\cite{li2024llava}       & 7B         & 1.0           & 13.3        & -           & 1.0           & 13.3        & -           \\
Llava-Rad~\cite{chaves2024towards} & 7B         & 15.4          & 30.6        & -           & 15.4          & 30.6        & -           \\
DCL~\cite{li2023dynamic}           & 0.25B      & 10.9          & 28.4        & -           & 10.9          & 28.4        & -           \\
VILA-M3-3B    & 3B         & 19.7          & 31.4        & 39.4           & 20.2          & 31.7        & 39.1           \\
VILA-M3-8B    & 8B         & 21.4          & 32.2        & 39.8           & 21.2          & 32.1        & 39.7        \\
VILA-M3-13B   & 13B        & 21.4          & 32.1        & 39.3           & 21.6          & 32.1        & 39.3           \\
VILA-M3-40B   & 40B        & \textbf{21.6}          & \textbf{32.2}        & 39.2        & \textbf{21.6}          & \textbf{32.2}        & 39.2        \\ \hline
\end{tabular}
\label{tab:report_generation_table}
\end{table*}

\subsection{Benchmarking}
\label{sec:benchmarking}
\noindent\textbf{Evaluation Metrics:} Accuracy as a measure was used to gauge the performance of close-ended questions for VQA tasks. F1-score was utilized for classification tasks. For report generation, we utilize three different metrics: BLEU-4, ROUGE, and GREEN score, which aims to be a more clinically relevant measure~\cite{ostmeier2024green}. In general, report generation is more challenging to evaluate and the metrics present different perspectives to understand the performance in a comprehensive manner.

\noindent\textbf{VILA-M3 Outperforms SOTA:} As a baseline comparison, we use the results from the Med-Gemini work directly \cite{yang2024advancing} and task-specific SOTA baselines \cite{li2024llava, li2023dynamic, chaves2024towards, Cohen2022xrv}. Overall, the results observed in Fig.~\ref{fig:radar_arch_plot} indicate superior performance to existing SOTA methods that are available from either open-source or closed-source. In particular, it can be seen that markedly higher performance can be achieved in 7 out of 8 metrics with much smaller models (a few billion parameters) compared to Med-Gemini, which contains 1.5 trillion parameters on this medical benchmark. 

Three models per parameter size variant were trained for 1, 2, and 3 epochs, respectively. 
The best performing VILA-M3 model was determined via comparison of all parameter size variants observed in Table.~\ref{tab:mainbenchmark_table}.

\noindent\textbf{VILA-M3 Benchmark:} 
We also evaluated VILA-M3 models on the VILA benchmark to ensure that they are not overfitting and degrading their baseline capabilities. Table.~\ref{tab:vila_benchmark} summarizes the results on the base VILA checkpoint and the VILA-M3 checkpoints after expert-guided IFT.

\subsection{Effectiveness of Expert-guided IFT}
Three key results prove the effectiveness of expert-guided IFT. The first is the observation that VILA-M3 outperforms SOTA models (Fig.\ref{fig:radar_arch_plot} \& Table.\ref{tab:mainbenchmark_table}). The second major observation is that the degradation on the VILA benchmark is not significant, after extensive training with expert-guided IFT, the performance of VILA-M3 only degrades by ~7\%, ~11\% and ~4\% for the 3B, 8B and 13B model variants (Table.~\ref{tab:vila_benchmark}). Contrastingly, the 40B model degrades by ~23\%, we attribute this to Llama models being superior than Yi \cite{young2024yi} models. The third observation can be drawn that VILA-M3 is not over-fitting to the expert IFT training process for epoch 1 and 2 models. This can be visually observed in Fig.~\ref{fig:heatmap} that over-fitting happens for epoch 3 models, these results are shown for the 8B model (results for more model variants are available in supplementary material). 
%\yin{single column? or wider/larger images into one fatter double column with less height?}
\begin{figure}[htbp]
    \centering
    \includegraphics[width=\columnwidth]{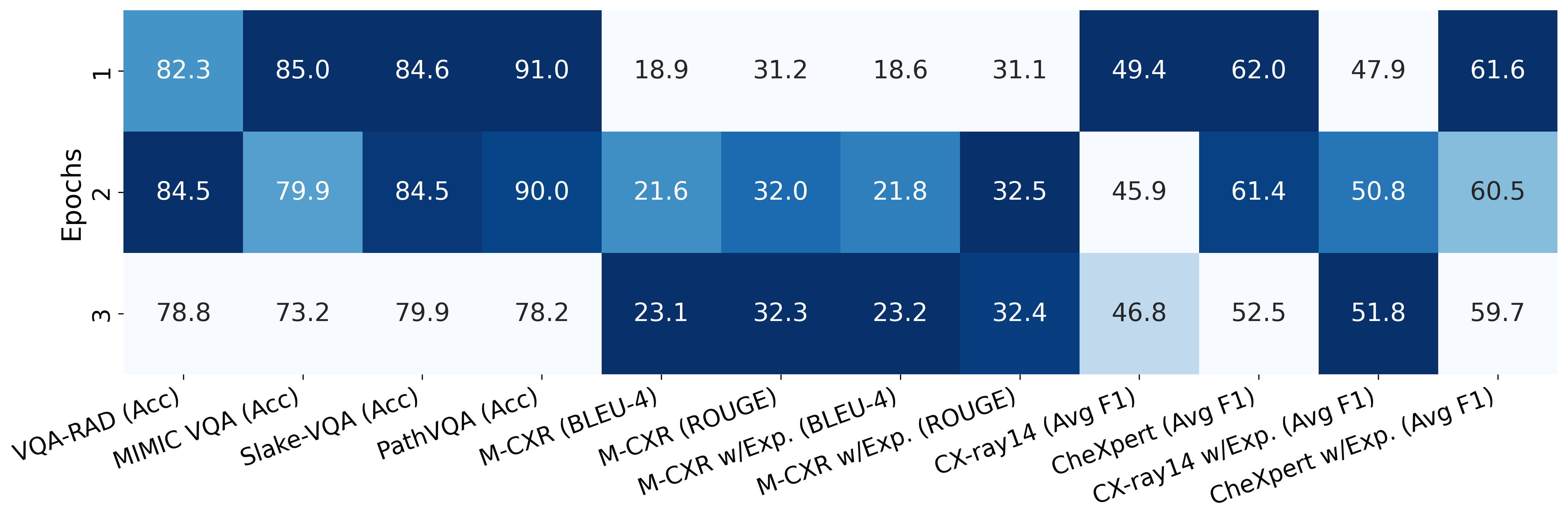}
    \caption{The heatmap shows the performance of the 8B model on all datasets with trained models at 1, 2, and 3 epochs. It can be observed that model performance degrades for the epoch 3 model.}
    \label{fig:heatmap}
\end{figure}
%%%%%%%%%%% additional classification results

\begin{table*}[htbp]
    \centering
    \caption{Performance comparison of different models on ChestX-ray14 classification tasks, both with and without expert models.}
    \scriptsize
    \resizebox{\textwidth}{!}{%
    \begin{tabular}{cc|ccc|ccc|cc}
    \hline
    \rowcolor{lightgray}
    \multicolumn{2}{c|}{Model} & \multicolumn{3}{c|}{ChestX-ray14 Classif. w/o Expert} & \multicolumn{3}{c|}{ChestX-ray14 Classif. \textit{With Expert}} & \multicolumn{2}{c}{Average} \\
    \hline
    \rowcolor{lightgray}
    Type & \# Params & Fracture & Pneumothorax & Lung opacity & Fracture & Pneumothorax & Lung opacity & w/o Expert & Expert \\
    \cline{1-10}
    GPT-4o & Unknown & 2.0 & 19.7 & 77.8 & - & - & - & 33.1 & - \\
    Med-Gemini & 1.5T & 5.5 & 55.3 & 79.9 & - & - & - & 46.9 & - \\
    TorchXrayVision & (ensemble) & - & - & - & 11.6 & 50.4 & 87.9 & - & 50.0 \\
    VILA-M3-3B & 3B & 7.0 & 54.5 & 83.8 & 10.8 & 55.2 & 88.1 & 48.4 & \textbf{51.3} \\
    VILA-M3-8B & 8B & 2.7 & 49.9 & 85.0 & 14.4 & 53.5 & 84.3 & 45.8 & \textbf{50.8} \\
    VILA-M3-13B & 13B & 7.6 & 55.9 & 86.2 & 13.6 & 51.5 & 88.5 & 49.9 & \textbf{51.2} \\
    VILA-M3-40B & 40B & 0.0 & 55.9 & 85.3 & 11.0 & 53.9 & 89.0 & 47.1 & \textbf{51.3} \\ \hline
    \end{tabular}%
    }
    \label{tab:classification_table_chestxray14_combined}
\end{table*}

\begin{table*}[htbp]
    \centering
    \caption{Performance comparison of VILA-M3 models on CheXpert classification tasks, with and without expert models.}
    \scriptsize
    \resizebox{\textwidth}{!}{%
    \begin{tabular}{cc|ccccc|ccccc|cc}
    \hline
    \rowcolor{lightgray}
    \multicolumn{2}{c|}{Model}  & \multicolumn{5}{c|}{CheXpert Classif. w/o Expert} & \multicolumn{5}{c}{CheXpert Classif. With Expert} & \multicolumn{2}{|c}{Average} \\ \cline{3-14}
    \rowcolor{lightgray}
    Type & \# Params   & Atel. & Cardio. & Consol. & Edema & Pl.Eff. & Atel. & Cardio. & Consol. & Edema & Pl.Eff. & w/o Expert & Expert \\ \hline
    GPT-4o & Unknown & 31.8 & 49.6 & 13.8 & 37.2 & 36.8 & - & - & - & - & - & 33.9 & - \\
    Med-Gemini & 1.5T & 49.7 & 72.0 & 23.0 & 32.7 & 64.4 & - & - & - & - & - & 48.4 & - \\
    TorchXrayVision & (ensemble) & - & - & - & - & - & 61.5 & 63.8 & 28.6 & 50.5 & 52.9 & - & 51.5 \\
    VILA-M3-3B & 3B & 59.0 & 69.3 & 24.0 & 71.8 & 62.9 & 61.2 & 71.5 & 35.0 & 63.7 & 72.4 & 57.4 & \textbf{60.7} \\
    VILA-M3-8B & 8B & 63.5 & 63.2 & 35.1 & 73.1 & 71.9 & 62.7 & 70.6 & 36.7 & 68.0 & 64.5 & 61.4 & \textbf{60.5} \\
    VILA-M3-13B & 13B & 61.9 & 71.1 & 20.5 & 65.4 & 60.2 & 62.8 & 76.0 & 39.4 & 67.1 & 62.5 & 55.8 & \textbf{61.5} \\
    VILA-M3-40B & 40B & 57.2 & 70.4 & 32.6 & 68.3 & 68.7 & 63.4 & 72.1 & 38.4 & 63.7 & 67.4 & 59.4 & \textbf{61.0} \\ \hline
    \end{tabular}%
    }
    \label{tab:classification_table_chexpert_combined}
\end{table*}

%%%%%%%%%%% end of additional classification results

\subsection{Ablation Studies}

\noindent\textbf{Expert Model Effect}:
To validate the findings of domain-expert models, we conduct experiments with VILA-M3 using the same training setting as outlined above in section \ref{sec:training_hyperparameters}. In particular, for the classification with and without expert feedback, we use the baselines of TorchXrayVision~\cite{Cohen2022xrv} and also compare it with the flagship model GPT-4o \cite{achiam2023gpt, hurst2024gpt} from OpenAI for both datasets ChestX-ray-14 and CheXpert, see Tables~\ref{tab:classification_table_chestxray14_combined} and \ref{tab:classification_table_chexpert_combined}. 

% Simiarly for report generation a study was conducted the and the baselines that were used for task-specific SOTA DCL, other generalist models are Llava-Med, Llava-Rad and  apart from Med-Gemini.
Experiments were performed for report generation with and without expert information inclusion (Table~\ref{tab:report_generation_table}). For comparison, we also utilize established baselines for the task-specific state-of-the-art (SOTA) ``Dynamic Graph Enhanced Contrastive Learning'' (DCL) model~\cite{li2023dynamic}. Notably, other baseline models considered in the study included Llama-Med~\cite{li2024llava}, Llama-Rad~\cite{chaves2024towards}, and Med-Gemini~\cite{yang2024advancing}, in addition to the DCL models~\cite{li2023dynamic}.

%Additionally we study the quantitative effect of question answering when a segmentation is provided for 2D images. \textcolor{red}{segmentation results pending}

For both classification and report generation tasks, the improvements achieved by inclusion of expert models can be observed in Tables \ref{tab:report_generation_table}, \ref{tab:classification_table_chestxray14_combined} \& \ref{tab:classification_table_chexpert_combined}. It can also be observed that large dataset models such as GPT-4o do not perform well on these tasks. 

For segmentation tasks, observing Fig. \ref{fig:qual_expert_seg} reveals that our model VILA-M3 fails to pick up on tumor findings if segmentation is not provided. However, with additional expert information included it is able to detect the finer features of a tumor.

\noindent\textbf{Balanced vs. Unbalanced Datasets}: The original size of datasets varies a lot, especially when the original numbers are compared between VQA datasets and MIMIC-CXR. Therefore, we increased the frequency of low-count datasets in a category-wise manner (based on VQA, expert, report gen., and classification type) to systematically balance the training dataset. The training settings were kept the same as outlined in section \ref{sec:training_hyperparameters}. The training dataset table with frequency counts can be found in the supplementary materials.

\begin{figure}[htbp]
    \centering
    \includegraphics[width=0.75\columnwidth]{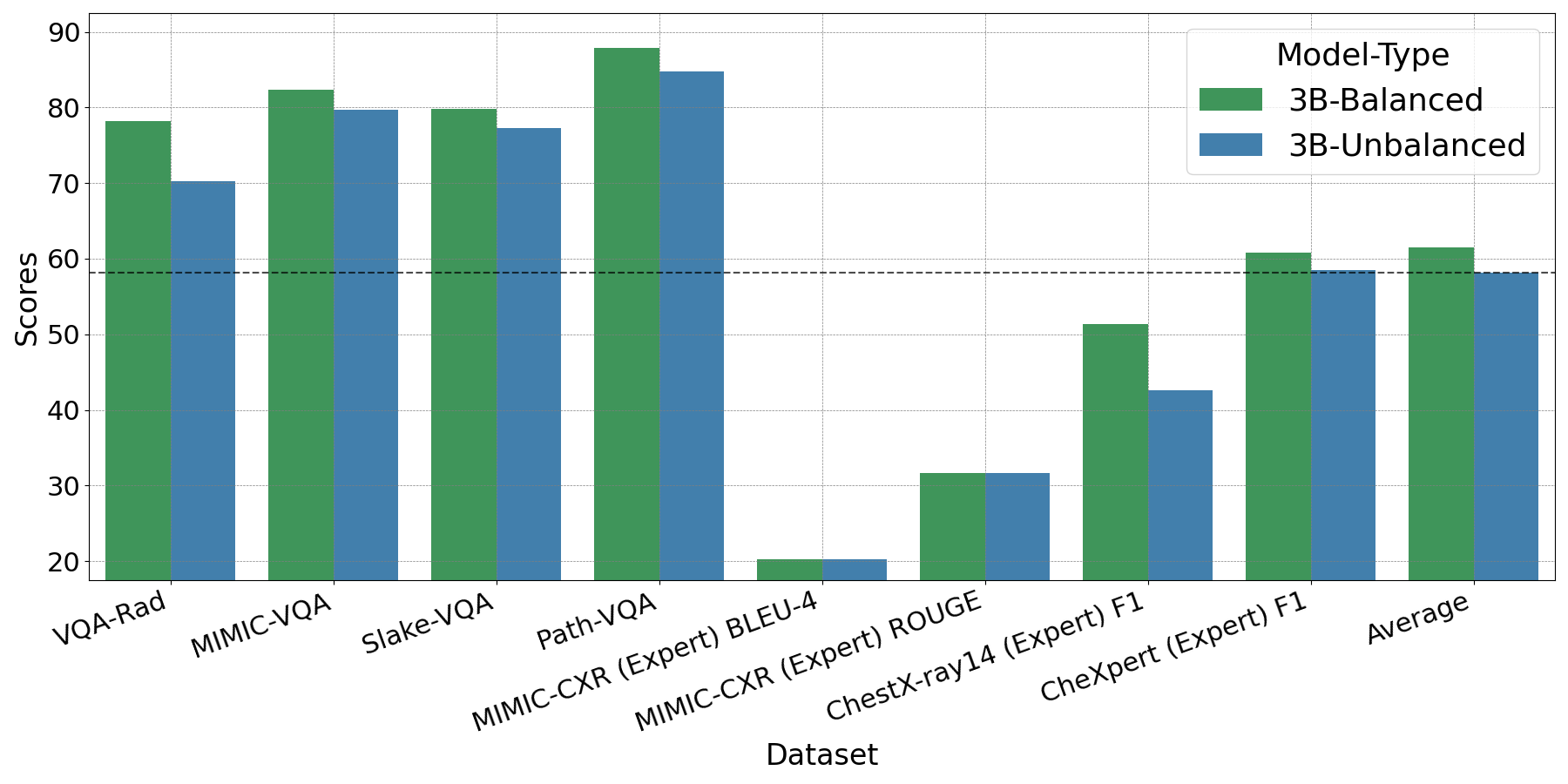}
    \caption{Comparison of VILA-M3 training with balanced and unbalanced healthcare datasets. Comparison for 3B model is shown with a training of two epochs each. %For exact data count of balancing please refer Table \ref{tab:balanced_training_dataset_stats}
    \label{fig:barchart}}
\end{figure}

The improvements based on the balanced training dataset as compared to the unbalanced dataset (original dataset size) can be observed in Fig.~\ref{fig:barchart}. Quantitatively, an average improvement of $\sim$4\% of all metrics is gained by balancing. 
%Furthermore, The training trends in Fig.~\ref{fig:unba_ba_trends} (supplementary) indicate that the balanced dataset provides an earlier convergence in the loss and an overall lower loss training trend is achieved.

\subsection{Training Convergence}
\begin{figure}[htbp]
    \centering
    \includegraphics[width=0.63\columnwidth]{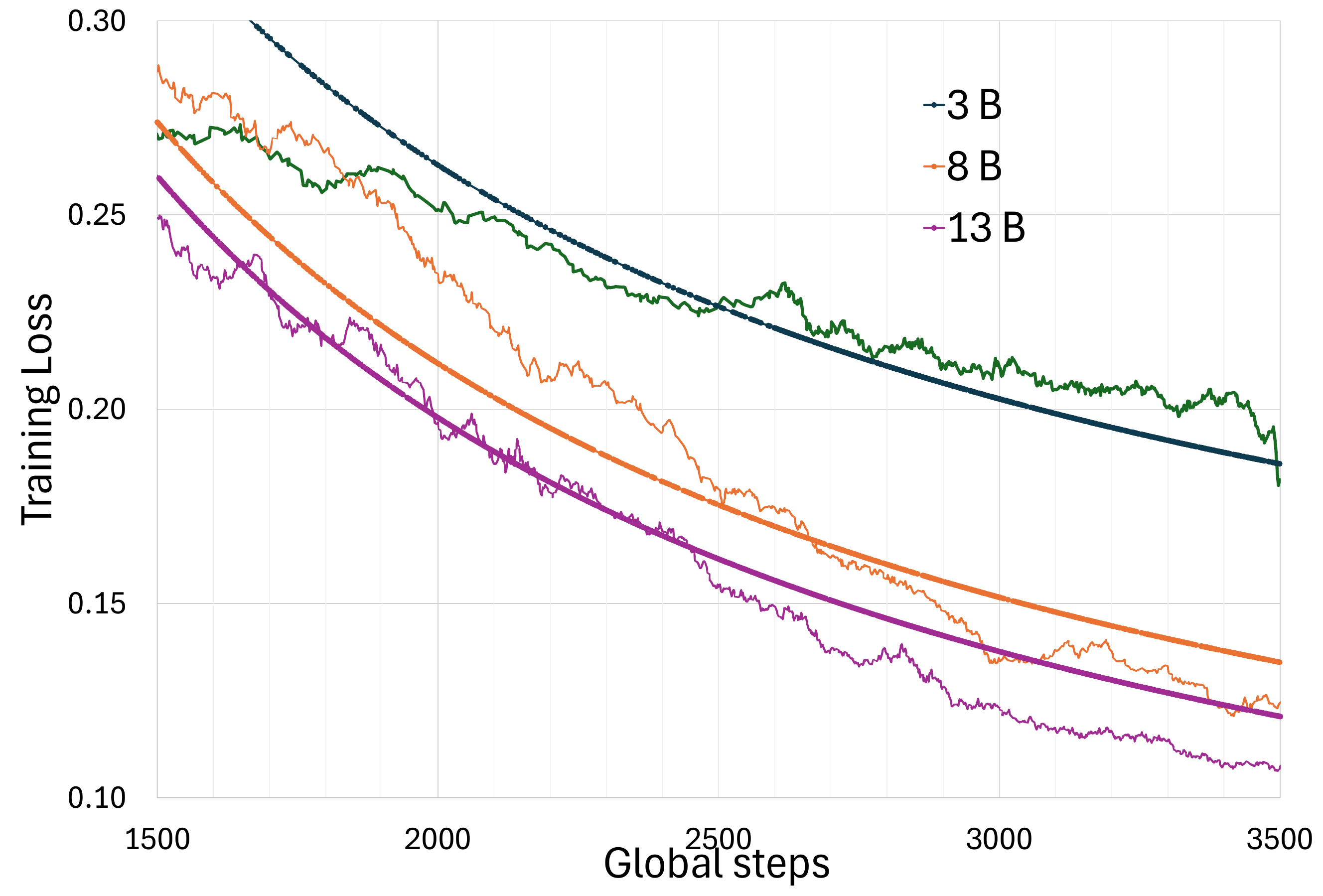}
    \caption{Comparison of training losses at different training steps for each model after an initial transient period.}
    \label{fig:loss_scaling_law}
\end{figure}

Despite different model architectures and sizes, we observe that the proposed training procedures converge successfully in all cases and are not sensitive to the choice of hyper-parameters. Taking the models trained with 4 nodes (32 GPUs) as an example, Fig.~\ref{fig:loss_scaling_law} shows that given the same training dataset, the training loss scales with model size as expected according to the empirical power law~\cite{kaplan2020scaling}.
%and Table~\ref{scaling_law_paramters} 

\section{Discussion \& Conclusion}

%\noindent\textbf{Discussion}: 
The proposed VILA-M3 model represents a significant advancement in multimodal machine learning for healthcare. It achieves SOTA performance as both a generalist model and one incorporating expert model responses to solve complex tasks. This capability allows the model to maintain high levels of performance when addressing general tasks while also leveraging expert knowledge to improve accuracy. % and efficacy in specialized areas 
With additional expert models, it can be seamlessly extended to various capabilities, such as registration, regression, etc.
Our work can be seen as a step towards enabling ``chain of thought'' capabilities in VLMs by letting the model reason about when to trigger an expert and how to incorporate its result in the final generated prediction.
For future work, we will explore the integration of Retrieval-Augmented Generation (RAG) to further enhance VILA-M3 by retrieving and incorporating relevant information from large datasets dynamically during inference. 
%This approach could improve the model's adaptability and responsiveness, particularly in applications requiring up-to-date information.
%
Additionally, we aim to expand VILA-M3 towards a multi-agent framework that directly incorporates smart expert models that can themselves decide whether to involve additional experts for consultation~\cite{guo2024large} to optimize performance across varied tasks. 
%
%%%%Agentic Framework, further improvements would be to make an agentic framework with expert models, however it requires careful design
%
%%%%However, implementing an agentic framework necessitates careful design to balance autonomy with control, ensuring the system remains reliable and interpretable. Addressing these design challenges will be crucial for advancing the model's capabilities and for its successful deployment in complex real-world applications.

%\noindent\textbf{Conclusion}: 
In conclusion, the proposed VILA-M3 framework can be used to train SOTA models that consider expert model responses to improve performance on particular tasks while simultaneously exhibiting good generalist performance. 
%%%The results of this work indicate that we need to carefully curate data and factor in expert information from the many medical domain-specialized models that have been trained in the past.

{
  \small
  \bibliographystyle{unsrt}
  \bibliography{main}
}
\clearpage
\section*{Supplementary Material}

\section{Dataset Details}

\subsection{Dataset Balancing}

As shown by experiments in the \textit{main} manuscript, dataset balancing is necessary to achieve the best performance when combining many datasets together (Main Manuscript, Figure 5). To balance the datasets, we categorized them into three broad categories: visual question answering (VQA), report generation, expert segmentation data, and language (see \textit{Category}, Table. \ref{tab:balanced_training_dataset_stats}). For example, if the dataset counts are summed up category-wise, their ratios for the entire dataset can be estimated. The proportion of each category differs significantly between the original and balanced versions. While VQA was modestly increased (from 24.9\% to 30.4\%), language and expert segmentation data were substantially increased (1.2\% to 5.5\% and 7.9\% to 34.8\%, respectively), whereas report generation was decreased (33.0\% to 14.7\%).
%%%%%%%%%%%%%% Dataset Tables

\begin{table}[h]
    \resizebox{\columnwidth}{!}{%
    \scriptsize
    \centering
    \begin{tabular}{l l l r r r}
    \toprule
    \rowcolor{lightgray}
    Type & Dataset & Category & Original & Freq. & Balanced \\
    \midrule
    Raw & USMLE & Lang & 10,178 & 10 & 101,780 \\
    Raw & RadVQA & VQA & 6,281 & 16 & 100,496 \\
    Raw & SLAKE & VQA & 5,972 & 16 & 95,552 \\
    Raw & PathVQA & VQA & 26,034 & 4 & 104,136 \\
    Expert & MIMIC-Diff-VQA & VQA & 129,232 & 2 & 258,464 \\
    Expert & MIMIC & Report & 270,000 & 1 & 270,000 \\
    Expert & VISTA3D & Seg & 50,000 & 8 & 400,000 \\
    Expert & BRATS & Seg & 15,000 & 16 & 240,000 \\
    \midrule
    & Total & & 819,456 & & 1,840,428 \\
    \bottomrule
    \end{tabular}}
    \caption{Balanced training dataset statistics showing original and balanced sample counts.}
    \label{tab:balanced_training_dataset_stats}
\end{table}

%%%%%%%%%%%%%% End of Dataset Tables

\subsection{Report Dataset Curation}
The preparation involves downloading datasets and refining report text with a Large Language Model (LLM) to create high-quality inputs for generating reliable medical reports. The primary dataset used is the \textbf{MIMIC Chest X-ray JPG Database v2.0.0} ~\cite{johnson2019mimiccxr,johnson2019mimic1cxrjpg}, which contains over 377,000 images and 227,827 free-text radiology reports. These data are de-identified to comply with HIPAA requirements. The process incorporates text enhancements and cleansing to optimize the quality of report inputs.

\subsubsection{Download Datasets}

The process begins with downloading the MIMIC-CXR-JPG dataset, which provides chest X-ray images and corresponding radiology reports. Data splits and labels are standardized, and enhanced text versions are utilized for improved report quality. The dataset is designed for tasks involving medical image analysis and natural language processing. Enhanced text versions ensure clarity and remove noise for better performance in model training. To refine the quality of the reports and eliminate noise, we utilize an enhanced text version developed by DCL~\cite{li2023dynamic}, and subsequently apply additional cleansing procedures to further optimize report accuracy.

\subsubsection{Sentence Pool Collection}

To standardize the language used in medical reports, a pool of sample sentences is created. This pool consists of commonly recurring phrases or sentence structures that appear in radiology reports. Using python scripts, the LLM (Llama-3.1-8b-instruct) analyzes the dataset and extracts these patterns, which are stored in a file (e.g., \texttt{sentence-pool.txt}) to guide the text transformation process.

\paragraph*{Example:}

\textbf{Sentence Pool:}

\begin{itemize}[noitemsep]
    \item The cardiac silhouette is normal in size.
    \item The lungs are low in volume.
    \item No pleural effusions.
    \item No pulmonary edema.
    \item There is mild pulmonary vascular congestion.
\end{itemize}

\textit{Prompt Example for Collection:}

\emph{"Analyze the medical report dataset and extract recurring sentence structures. Compile these into a pool of sample sentences to be used for standardizing input text."}

\subsubsection{Text Conversion Using LLM}

The LLM (Llama-3.1-8b-instruct) utilizes the sentence pool to process the text, replacing or reformatting sentences and medical terminology into a consistent and standardized format. This step ensures uniform input data for the VLM model, improving its ability to generate reliable and accurate reports.

\paragraph*{Example:}

\begin{itemize}[noitemsep]
    \item \textbf{Before Conversion:}

    "Lungs are low in volume. Congestion of the pulmonary vasculature, small bilateral pleural effusions and presence of septal lines reflect mild pulmonary edema."
    
    \item \textbf{After Conversion:}

    "The lungs are low in volume. There is mild pulmonary vascular congestion. No pleural effusions are seen."
\end{itemize}

\textbf{Prompt Example for Conversion:}

\emph{"Using the provided sentence pool, transform the medical report text into a consistent format. Ensure the output text aligns with the predefined sentence structures without altering the content's meaning. Return only the standardized text."}

\subsection{Expert Dataset Curation}

\textbf{Segmentation:} For segmentation models, we can train VILA-M3 to trigger a suitable expert model together with several different input arguments. For example, when triggering the VISTA3D model, we generate several variations (using Llama-3.1) of instructions to the model to segment a certain anatomical structure in the image and give a suitable trigger message. For instance:
\begin{itemize}[align=parleft]
\scriptsize
    \item User: \texttt{Can you identify any liver masses or tumors?}
    \item VILA-M3: \texttt{This looks like a CT image. Let me trigger <VISTA3D(hepatic tumor)>.} 
\end{itemize}

\begin{itemize}[align=parleft]
\scriptsize    
    \item User: \texttt{Can you assist me in segmenting the bony structures in this image?}
    \item VILA-M3: \texttt{I segmented the skeleton using <VISTA3D(skeleton)>.}
\end{itemize}

\begin{figure*}[h]
    \centering
  \begin{subfigure}{0.2\textwidth}
    \includegraphics[width=\textwidth]{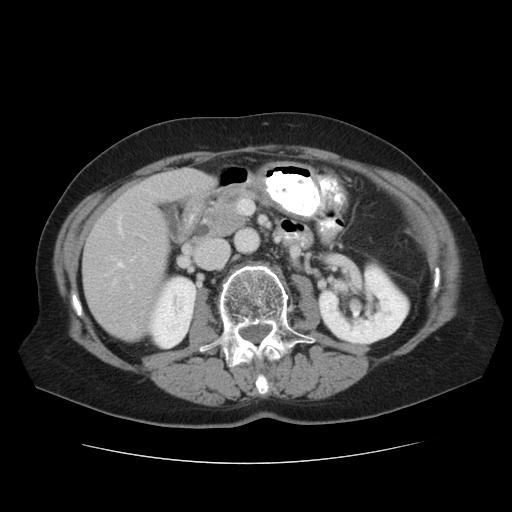}
    \caption{original image}
  \end{subfigure}
  \begin{subfigure}{0.2\textwidth}    
    \includegraphics[width=\textwidth]{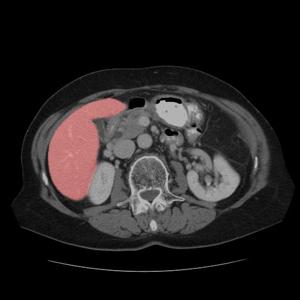}
    \caption{\texttt{hepatic tumor}}
  \end{subfigure}    
  \begin{subfigure}{0.2\textwidth}    
    \includegraphics[width=\textwidth]{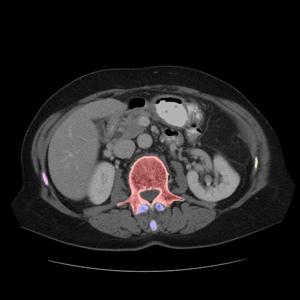}    
    \caption{\texttt{skeleton}}
  \end{subfigure}
  \begin{subfigure}{0.2\textwidth}    
    \includegraphics[width=\textwidth]{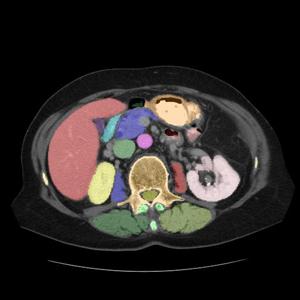}
    \caption{\texttt{everything}}    
  \end{subfigure}
\caption{(a) The original CT image slice. (b-d) The selected argument by VILA-M3 to the VISTA3D expert model call. \label{fig:segment_args}}
\end{figure*}

\begin{itemize}[align=parleft]
\scriptsize
    \item User: \texttt{Segment the entire image.}
    \item VILA-M3: \texttt{I segmented the entire image using <VISTA3D(everything)>.}    
\end{itemize}
For these types of instructions, VILA-M3 predicts the suitable arguments for the VISTA3D model in order to trigger the segmentation of the correct anatomical structure as illustrated in Fig.~\ref{fig:segment_args}. We built these types of instruction and answer pairs into the expert training dataset in order to build segmentation capabilities into VILA-M3. 
Table~\ref{tab:expert_data} summarizes the expert models and datasets used in this work to build up the expert training dataset.

\begin{table}[htbp]
\caption{Expert Selection Training Data. \label{tab:expert_data}}
\resizebox{\columnwidth}{!}{%
\footnotesize
\begin{tabular}{llp{11em}}
\rowcolor[HTML]{F2F2F2} 
\toprule
\textbf{Modality}    & \textbf{Expert}            & \textbf{Datasets}                                          \\
\toprule
CT          & VISTA3D           & MSD (liver, spleen, pancreas), TotalSegmentatorV2 \\
MRI         & BRATS (SegResNet) & BRATS (2018)                                      \\
CXR & TorchXRayVision   & MIMIC (Reports, VQA)        
%\bottomrule
\end{tabular}}
\end{table}

\noindent
\textbf{Report Generation:}
Converting expert model predictions into conversation format involves transforming classification outputs into structured dialogues for AI training in medical report generation. Initially, ensemble predictions are created by combining probabilities of various medical conditions (like Atelectasis and Effusion) from multiple expert models from TorchXRayVision applied to chest X-ray images. These probabilities are interpreted to determine the presence (\textquotedblleft yes\textquotedblright) or absence (\textquotedblleft no\textquotedblright) of each condition based on a threshold.

The formatted expert predictions are then integrated into conversation prompts that include an image placeholder, a prompt for report generation, and the expert results as additional information. For example, the prompt might be:

\begin{quote}
\texttt{\textless image\textgreater\\
Describe the image in detail.\\
When answering, please incorporate the expert model results:\\
Atelectasis: yes\\
Cardiomegaly: no\\
Effusion: yes}
\end{quote}

Each conversation is assembled into a structured format containing a unique identifier, the image reference, and the dialogue between the human and the VLM model. This approach enriches the dataset with various prompt types and simulates interactions where the VLM model provides diagnoses based on image analysis and expert insights, enhancing the model's ability to generate accurate and contextually rich medical reports.

\label{sec:dataset}

\section{Training \& Compute Details}
\subsection{Training Details}

\begin{figure}[h]
    \centering
    \includegraphics[width=\columnwidth]{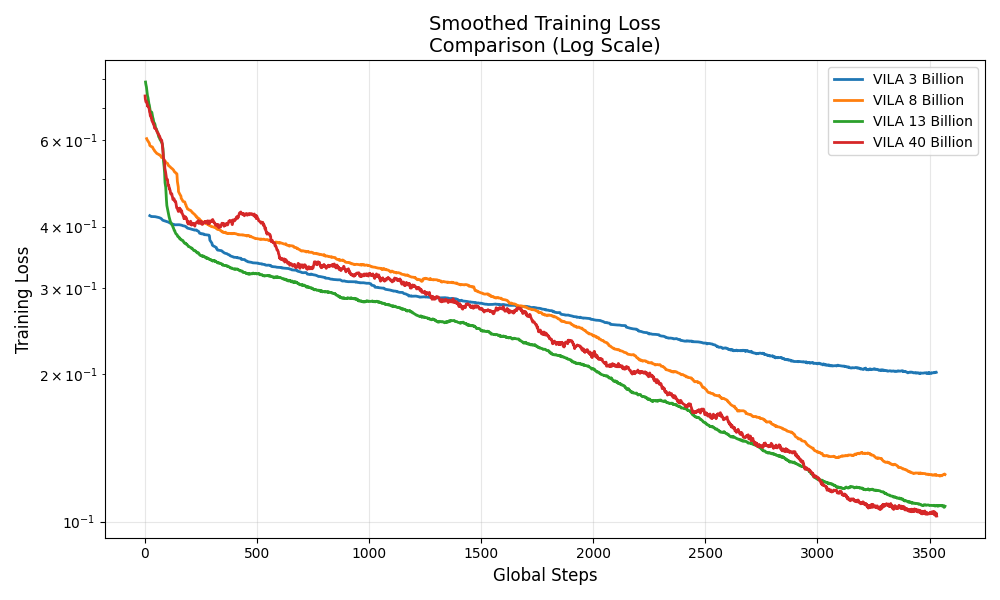}
    \caption{Comparison of training trends across all model variants by parameter size. The compared variants are 3, 8, 13, and 40 billion parameter models.}
    \label{fig:training_trends}
\end{figure}

The loss curves for all model variants when training for 2 epochs can be observed in Fig.~\ref{fig:training_trends}. The largest change in the training loss can be observed between 3B and 8B, 13B and 40B models. There is a significant learning gap that one can observe between 3B and 8B onwards, which demonstrates that the 3B model learning capabilities are limited. It should also be noted that the 40B model has a noisier training curve as compared to all other models, this can be attributed to two major factors, the first being that the vision backbone is much larger (6B parameters) as compared to other configurations and the second factor being that it is a Yi model \cite{young2024yi} and their learning behavior is different. The training trends overall also indicate that larger models have diminishing returns in terms of loss convergence, note that the 8B, 13B and 40B are quite close in terms of the final loss values. The 40B model loss stays quite high during training and takes a longer time to converge below the loss values of other models, indicating that it is slower to train.

\noindent
\subsection{Inference Compute}
We have successfully deployed an inference workflow of the proposed framework.
While there are additional techniques (such as TensorRT) could be used to improve the system throughput,
we list the computational costs without `bells and whistles' to show the technical feasibility of practical usages in Table~\ref{tab:computational_cost}.
%%%%%%%%%%% computational cost
\begin{table}[h]
    \centering
    \scriptsize
    \resizebox{\columnwidth}{!}{%
    \begin{tabular}{cc|ccc}
    \hline
    \rowcolor{lightgray}
    \multicolumn{2}{c|}{Model} & \multicolumn{3}{c}{VLM Properties} \\ \hline
    Type & \# Params & VRAM (GB) & tokens/sec & Max context len \\ \hline
    % Row 1 (example, fill in values accordingly)
    VILA-M3 & 3B & 7 & 41 & 4,096 \\
    VILA-M3 & 8B & 18 & 32 & 4,096 \\
    VILA-M3 & 13B & 30 & 26 & 4,096  \\
    VILA-M3 & 40B & 77 & 9 & 4,096  \\
    \end{tabular}}
    \caption{Inference computational costs for VILA-M3 variants.}
    \label{tab:computational_cost}
\end{table}
%%%%%%%%%%%%%%%%%%%% end of computational cost

\section{Additional Experiments}

\subsection{Scaling Law Experiment Additional Details}

%~\ref{fig:loss_scaling_law}
\begin{table}[h]
    \scriptsize
    \centering
    \caption{Fits to $L(N, S)$ of the loss scaling law defined in [25] Equation~5.6. These parameters are visualized in Main Manuscript Fig~6.}
    \begin{tabular}{c|cccc}
        \hline
        \rowcolor{lightgray}
        \textbf{Parameter} & $\alpha_N$ & $\alpha_S$ & $N_c$ & $S_c$ \\
        \hline
        \textbf{Value} & 0.78 & 1.09 & $1.50 \times 10^{8}$ & $3.92 \times 10^{2}$ \\
    \end{tabular}
    \label{scaling_law_paramters}
\end{table}

In the Main Manuscript, Fig.~6 shows that after an initial warm-up period, the learning curve of the fine-tuning processes can be approximately fit by a universal function parameterized by model size and number of steps (using 32 GPUs in this case).  Table~\ref{scaling_law_paramters} summarizes the empirical fit of the power law parameters based on Equation~5.6 in ~\cite{kaplan2020scaling}.  It is interesting to observe that although our fine-tuning dataset size is relatively small compared with a typical LLM training scratch setup, the training process scales according to the model size and training steps following a similar pattern.

\subsection{Balanced \& Unbalanced Datasets Training}

\begin{figure*}[h]
    \centering
    \includegraphics[width=0.9\textwidth]{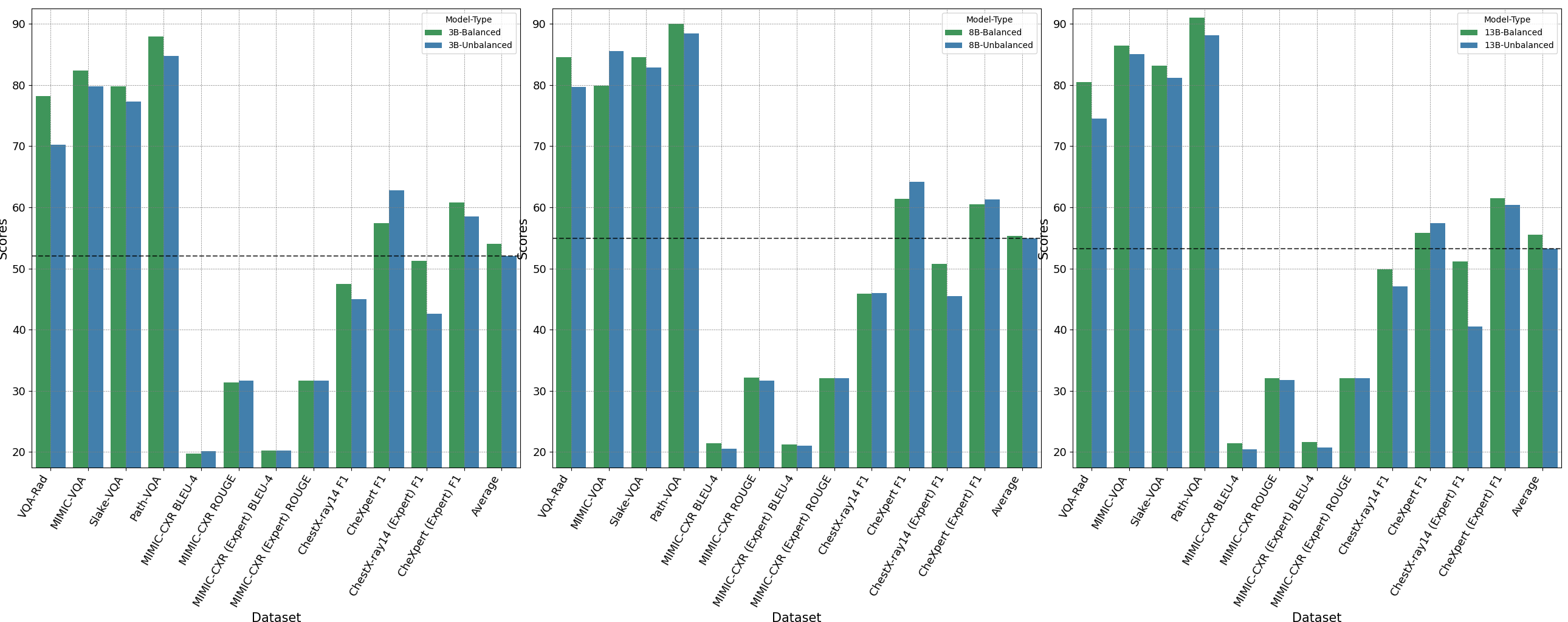}
    \caption{Comparison of VILA-M3 training with balanced and unbalanced healthcare datasets. Comparison for 3B, 8B and 13B model are shown with a training of two epochs each}
    \label{fig:balanced_unbalanced_all_models}
\end{figure*}

Since the original size of datasets varies a lot as can be observed in Table.~\ref{tab:balanced_training_dataset_stats}. In the main manuscript we showed the comparison between balanced and unbalanced datasets for the 3B model. Here, we plot additional results for the 8B and 13B models both as shown in Fig.\ref{fig:balanced_unbalanced_all_models}. 

\begin{figure}[h]
    \centering
    \includegraphics[width=0.8\columnwidth]{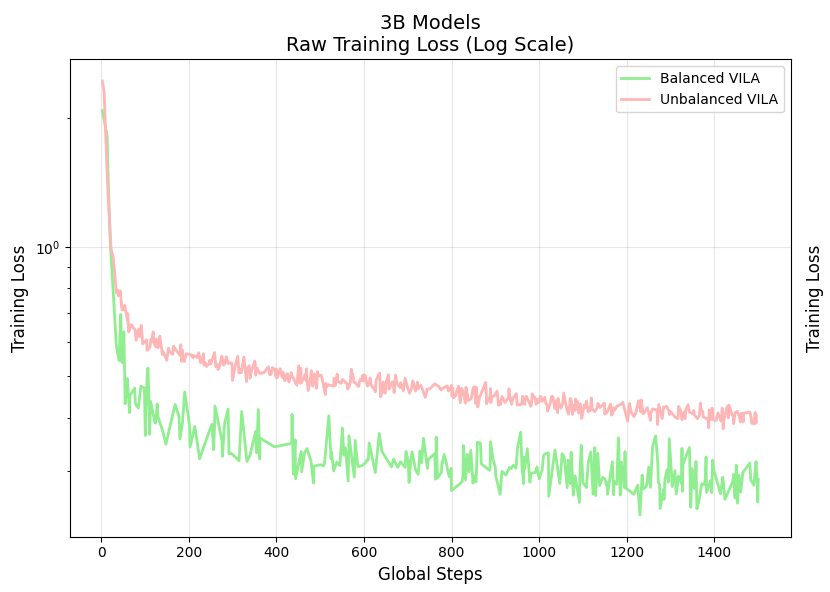}
    \caption{Comparison of training trends across 3B model to show the effect of balanced and unbalanced dataset on model training.}
    \label{fig:unba_ba_trends}
\end{figure}

The improvements based on the balanced training dataset as compared to the unbalanced dataset (original dataset size) can be observed in Fig.~\ref{fig:balanced_unbalanced_all_models}. Quantitatively, an average improvement of $\sim$4\% of all metrics is gained by data balancing for the 3B and 13B models. The 8B model shows an improvement of $\sim$2\%. 
Furthermore, the training trends in Fig.~\ref{fig:unba_ba_trends} indicate that the balanced dataset provides an earlier convergence in the loss and an overall lower loss training trend is achieved.

\subsection{Comparisons With GPT-4o For Classification}

Tables 5 \& 6 (Main Manuscript) show a comparison with GPT-4o \cite{achiam2023gpt, hurst2024gpt} for classification experiments. The inference on GPT-4o was performed with the same prompt as being used for VILA-M3. The images were pre-appended to the prompt via API call using a python script and then the responses were collected with retry attempts set to 10. 

We often found that with API usage the GPT-4o provided inconsistent responses as it took more 3-5 retries for more than quite a few images for both CheXpert and Chest X-ray datasets. We also found that if we appended the \textit{expert model information}, the GPT-4o refused to provide responses for many images and therefore a quantitative analysis would be unfair. For more than 50\% images responses could not be successfully retrieved. We believe the guardrails of GPT-4o likely come into effect when a sufficient amount of medical terminology is used within the prompt itself.

\subsection{Statistical significances of the classification experiments}
We conduct McNemar's Chi-square tests to show that the proposed M3 models (with expert models) significantly differ from GPT-4o (results shown in Main Manuscript Table~5 and 6).  The values that are smaller than $0.05$ indicate significance. For the corresponding model and specific class denoted with \textbf{*}, M3 is significantly better than GPT-4o.

\begin{table}[h!]
\scriptsize
\centering
\begin{tabular}{c|ccc}
\hline
\rowcolor{lightgray}
\textbf{}    & Fracture & Pneumothorax & Lung opacity \\ \hline
M3-3B  & 5.5e-10$^{*}$ & 3.2e-5$^{*}$ & 6.1e-41$^{*}$ \\
M3-8B  & 1.3e-1    & 3.1e-1   & 4.5e-14$^{*}$ \\
M3-13B & 5.9e-20$^{*}$ & 5.0e-8$^{*}$ & 1.0e-44$^{*}$ \\
M3-40B & 3.1e-12$^{*}$ & 2.5e-8$^{*}$ & 3.2e-49$^{*}$ \\
\end{tabular}
\caption{$p$-value for the null hypothesis that the classification models perform at the same level as GPT-4o for each class on ChestXray14.}
\label{p-value-chestxray14}
\end{table}

\begin{table}[h]
\scriptsize
\resizebox{\columnwidth}{!}{%
\centering
\begin{tabular}{c|ccccc}
\hline
\rowcolor{lightgray}
 & Atel. & Cardio. & Consol. & Edema & Pl. Eff.
\\ \hline
M3-3B   & 2.3e-53$^{*}$ & 2.1e-37$^{*}$ & 6.5e-43$^{*}$ & 7.0e-48$^{*}$ & 2.6e-51$^{*}$ \\ 
M3-8B   & 3.7e-59$^{*}$ & 1.0e-16$^{*}$ & 3.4e-33$^{*}$ & 1.3e-30$^{*}$ & 1.3e-32$^{*}$ \\ 
M3-13B  & 2.0e-46$^{*}$ & 2.9e-30$^{*}$ & 4.6e-22$^{*}$ & 1.3e-41$^{*}$ & 1.3e-32$^{*}$ \\ 
M3-40B  & 6.6e-52$^{*}$ & 2.5e-18$^{*}$ & 2.8e-23$^{*}$ & 3.2e-47$^{*}$ & 3.8e-40$^{*}$ \\ 
\end{tabular}}
\caption{$p$-value for the null hypothesis that the classification models perform at the same level as GPT-4o for each class on CheXpert.}
\label{p-value-chexpert}
\end{table}

\subsection{Additional Results Expert-Guided IFT}

From the main manuscript in Fig 4. it was observed that the 8B model overfits when the 3 epoch variant is trained. In Fig. \ref{fig:heatmap} results for all datasets for all model variants (3B, 8B, 13B and 40B) can be observed. The performance deterioration for all datasets is evident for the 3 epoch model for all variants with the exception of report generation tasks. Since the report generation task is of a much more complex nature the models do not overfit to it.

\begin{figure*}[h]
    \centering
    \includegraphics[width=\textwidth]{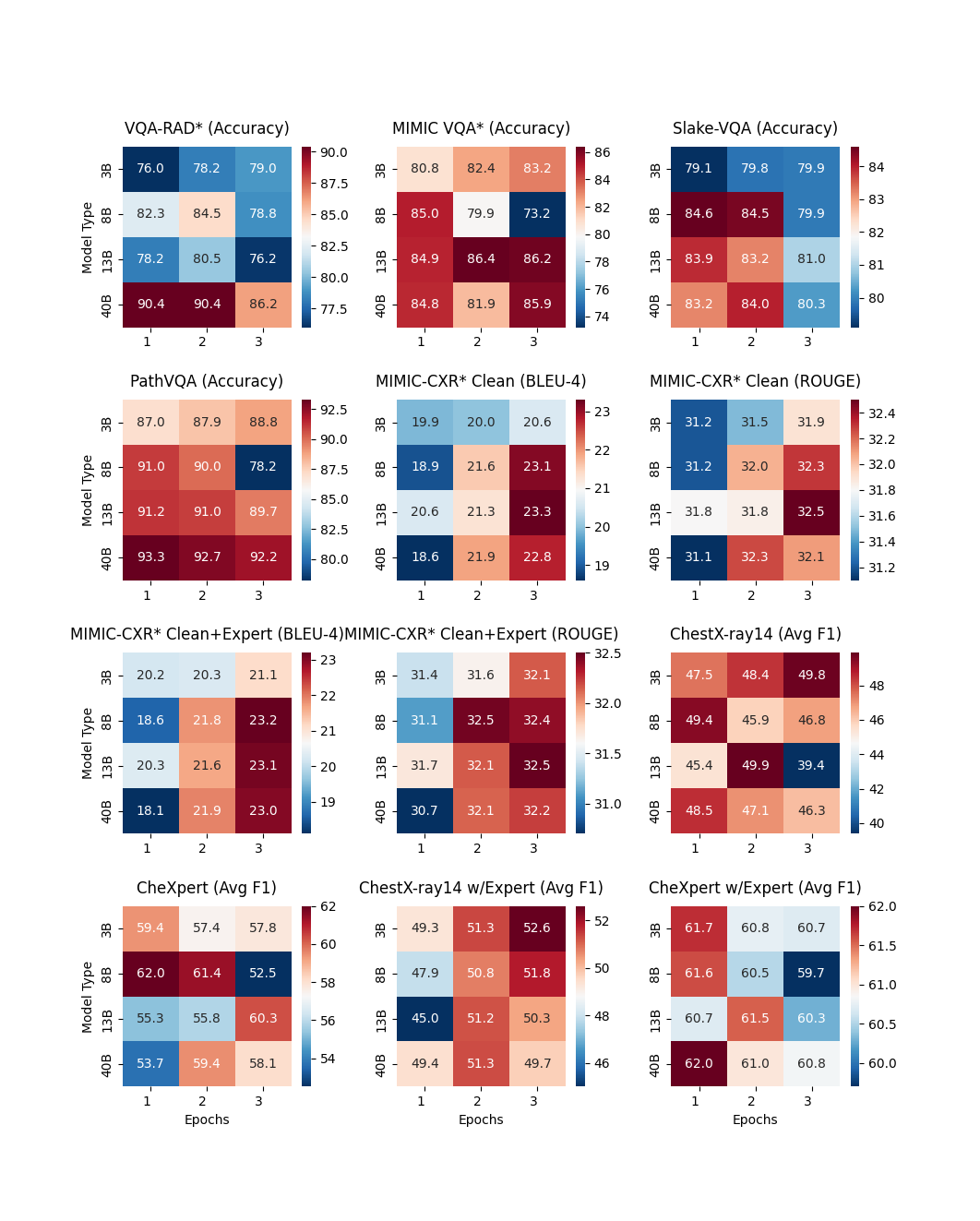}
    \caption{The heatmap shows the performance of the all model variants 3B, 8B, 13B and 40B on all datasets with trained models at 1, 2 and 3 epochs.}
    \label{fig:heatmap}
\end{figure*}

\subsection{Additional Results on Report Generation}

We conducted a detailed analysis of the test set results for report generation, focusing on the disease and pattern distribution (multi-label ground truth) from the \textbf{MIMIC Chest X-ray JPG Database v2.0.0}. This evaluation utilized the best-performing 40B model, covering both with and without the integration of expert model predictions.
Table~\ref{fig:detailed-report} shows performance metrics with and without expert model predictions across various categories of medical findings. Key metrics include \texttt{BLEU-4}, \texttt{ROUGE}, and GREEN score (\texttt{GREEN}), evaluated for both settings. For most categories, the inclusion of expert models resulted in marginal improvements in the \texttt{GREEN} metric, which measures overall accuracy. For instance, in the \textit{Atelectasis} category, \texttt{GREEN} improved from 35.93 to 36.56, and in \textit{Cardiomegaly}, it increased from 38.99 to 39.79. However, some metrics, such as \texttt{BLEU-4} in categories like \textit{Fracture} and \textit{Lung Lesion}, showed slight decreases when expert model predictions were used.

Notably, the \textit{No Finding} category exhibited the largest improvement in \texttt{green} (45.60 to 46.57), indicating that expert models may be particularly effective in identifying cases with no abnormalities. These results suggest that while the integration of expert models generally enhances certain metrics, the gains are context-dependent and may vary across different medical conditions.

\begin{table*}[ht]
\centering
\caption{Detailed performance on report generation with and without expert models.}
\resizebox{\textwidth}{!}{%
\begin{tabular}{llccccccc}
\toprule
\multirow{2}{*}{\textbf{Category}} & \multirow{2}{*}{\textbf{Quantity}} & \multicolumn{3}{c}{\textbf{Without Expert}} & \multicolumn{3}{c}{\textbf{With Expert}} \\
\cmidrule(lr){3-5} \cmidrule(lr){6-8}
 & & \textbf{BLEU-4 (↑)} & \textbf{ROUGE} & \textbf{GREEN} & \textbf{BLEU-4} & \textbf{ROUGE} & \textbf{GREEN} \\
\midrule
Atelectasis & 494 & 19.21 & 31.32 & 35.93 & 19.03 & 31.31 & 36.56 \\
Cardiomegaly & 426 & 18.62 & 32.11 & 38.99 & 18.38 & 31.98 & 39.79 \\
Consolidation & 99 & 18.46 & 31.47 & 37.23 & 18.20 & 30.70 & 34.97 \\
Edema & 451 & 20.69 & 33.39 & 38.97 & 20.54 & 33.68 & 39.66 \\
Enlarged Cardiomediastinum & 70 & 16.10 & 27.63 & 31.86 & 16.67 & 28.57 & 35.28 \\
Fracture & 53 & 19.12 & 31.07 & 37.51 & 17.24 & 30.20 & 35.88 \\
Lung Lesion & 83 & 20.75 & 32.92 & 38.89 & 18.64 & 31.56 & 36.17 \\
Lung Opacity & 759 & 18.60 & 30.35 & 35.92 & 17.96 & 30.18 & 35.30 \\
No Finding & 561 & 21.20 & 34.25 & 45.60 & 21.81 & 34.16 & 46.57 \\
Pleural Effusion & 678 & 18.52 & 30.60 & 34.53 & 18.20 & 30.44 & 35.12 \\
Pleural Other & 37 & 19.27 & 29.37 & 32.09 & 17.32 & 27.42 & 36.10 \\
Pneumonia & 219 & 19.39 & 32.30 & 38.41 & 20.14 & 32.91 & 39.43 \\
Pneumothorax & 58 & 18.76 & 29.97 & 30.02 & 17.39 & 28.62 & 27.71 \\
Support Devices & 635 & 19.25 & 30.70 & 35.24 & 19.19 & 30.42 & 35.91 \\
\bottomrule
\end{tabular}}
\label{fig:detailed-report}
\end{table*}

\section{Additional Discussion}

As also stated in the main manuscript, we will explore the integration of Retrieval-Augmented Generation (RAG) to further enhance VILA-M3 by retrieving and incorporating relevant information from large datasets dynamically during inference. 

%This approach could improve the model's adaptability and responsiveness, particularly in applications requiring up-to-date information.
%Additionally, we aim to expand VILA-M3 towards a multi-agent framework that directly incorporates smart expert models that can themselves decide whether to involve additional experts for consultation~\cite{guo2024large} to optimize performance across varied tasks.  

Further improvements would be to make an agentic framework with expert models, however it requires careful design. However, implementing an agentic framework necessitates careful design to balance autonomy with control, ensuring the system remains reliable and interpretable. Addressing these design challenges will be crucial for advancing the model's capabilities and for its successful deployment in complex real-world applications.

The results of this work indicate that we need to carefully curate data and factor in expert information from the many medical domain-specialized models that have been trained in the past.

\end{document}